\documentclass[preprint,12pt]{elsarticle}

\usepackage[utf8]{inputenc}
\usepackage[T1]{fontenc}

\usepackage{amssymb}

\usepackage[usenames,dvipsnames]{color} 
\usepackage{soul} 

\usepackage{amsmath, amsfonts, bm} 
\usepackage{bbm} 
\usepackage{stmaryrd} 
\usepackage{bbm} 

\usepackage{float} 
\usepackage{dirtytalk} 

\usepackage[ruled]{algorithm2e}
\SetKwComment{Comment}{$\triangleright$\ }{}

\usepackage{subcaption}

\DeclareMathOperator*{\argmax}{arg\,max} 
\DeclareMathOperator*{\argmin}{arg\,min} 
\DeclareMathOperator*{\diag}{\mathrm{diag}} 

\newtheorem{assumption}{P} 

\usepackage{prettyref}
\newrefformat{sec}{Section~\ref{#1}} 
\newrefformat{eq}{Equation~\ref{#1}} 
\newrefformat{fig}{Figure~\ref{#1}} 
\newrefformat{th}{Theorem~\ref{#1}} 
\newrefformat{prop}{Proposition~\ref{#1}} 
\newrefformat{P}{\textbf{P\ref{#1}}} 

\begin{document}

\begin{frontmatter}

\title{Unsupervised Complex Semi-Binary Matrix Factorization for Activation Sequence Recovery of Quasi-Stationary Sources}


\author[inst1]{Romain Delabeye\corref{c1}}
\author[inst1]{Martin Ghienne}
\author[inst1]{Olivia Penas}
\author[inst1]{Jean-Luc Dion}

\cortext[c1]{Corresponding author}
\affiliation[inst1]{
            organization={Quartz Laboratory, EA7393, ISAE-Supméca}, 
            addressline={3 rue Fernand Hainaut}, 
            city={Saint-Ouen},
            postcode={93400}, 
            country={France}
            }

\begin{abstract}

Advocating for a sustainable, resilient and human-centric industry, the three pillars of Industry 5.0 call for an increased understanding of industrial processes and manufacturing systems, as well as their energy sustainability.
One of the most fundamental elements of comprehension is knowing when the systems are operated, as this is key to locating energy intensive subsystems and operations.
Such knowledge is often lacking in practice.
Activation statuses can be recovered from sensor data though.
Some non-intrusive sensors (accelerometers, current sensors, etc.) acquire mixed signals containing information about multiple actuators at once.
Despite their low cost as regards the fleet of systems they monitor, additional signal processing is required to extract the individual activation sequences.
To that end, sparse regression techniques can extract leading dynamics in sequential data. 
Notorious dictionary learning algorithms have proven effective in this regard.
This paper considers different industrial settings in which the identification of binary sub-system activation sequences is sought.
In this context, it is assumed that each sensor measures an extensive physical property, source signals are periodic, quasi-stationary and independent, albeit these signals may be correlated and their noise distribution is arbitrary.
Existing methods either restrict these assumptions, e.g., by imposing orthogonality or noise characteristics, or lift them using additional assumptions, typically using nonlinear transforms.

This paper addresses these limitations, and introduces the unsupervised complex semi-binary matrix factorization ($\mathbb{C}$SBMF) as its main contribution.
In particular, we show that the exact recovery of source activation sequences from non-intrusive sensor data is intrinsically tied to the presence of problematic phase shifts, the causes of which are detailed.
A greedy algorithm is proposed, iteratively resynchronizing sources to converge towards the maximum decomposition of each operation despite these phase shifts.
The $\mathbb{C}$SBMF is verified and compared to existing techniques on synthetic use cases, then validated on experimental data with signals of different nature.
To that occasion, the CAFFEINE dataset for unsupervised time series multi-label classification is introduced.

\end{abstract}


\end{frontmatter}




\section{Introduction}
\label{sec:intro}

The manufacturing industry is inherently energy-intensive, accounting for around 37\% of global final energy consumption in 2022 \cite{iea2022world}.
Amid the recent energy crisis, some factory pilots are looking to increase the efficiency of their machines and processes, aiming for substantial reductions in energy use and associated carbon emissions.
Among the many solutions sought to achieve this objective, digital twins stand out, a popular cross-industry concept that has seen a rapid rise in recent years.
A digital twin is \textit{\say{a set of adaptive models that emulate the behavior of a physical system in a virtual system getting real time data to update itself along its life cycle. The digital twin replicates the physical system to predict failures and opportunities for changing, to prescribe real time actions for optimizing and/or mitigating unexpected events observing and evaluating the operating profile system}} \cite{semeraro2021digital}.
In particular, this enables a number of enhancements, from predicting energy consumption, locating energy drifts induced by faulty components, misuse or environmental changes, to optimizing machine control, component replacement, and process scheduling.

These improvements come at a cost though.
A manufacturing plant is characterized by the many cyber-physical systems (CPS) it contains and the similarities they may present.
If scalability is sought as regards production, monitoring is no exception.
That is, in number of cases, few sensors must monitor a large fleet of machines.
Not only are they able to monitor multiple systems at once, but their implementation does not require direct intervention on these systems' hardware or software.
Placing sensors inside a machine, provided that a suitable location can be found without tampering with the production system, often requires to stop the said machine during the intervention.
Intrusive sensors may also require rewiring, reprogramming or accessing data from a CPS's dedicated programmable logic controller (PLC).
Actuator-specific operating statuses may also be programmed in a low-level language without being returned to the user interface.
Overall, the activation sequences are difficult to retrieve.
Manually labeling data as a post-processing layer is expensive at plant scale, and so is the development of a physical model for each actuator.
As an alternative, with a view to learn energy consumption models, estimate and predict actuator-specific performance indicators, the activation sequences can be extracted from sensor data instead.
Hence, in order to unlock the above-mentioned applications, this paper focuses on fully unsupervised non-intrusive load monitoring (NILM), and more specifically on the recovery of actuator activation sequences in sensor data.
Moreover, the elements presented here are not limited to current load, but to piecewise mixed quasi-stationary periodic signals in the broadest sense.

Unsupervised NILM, or energy disaggregation, aims to discover the active appliances in energy consumption data \cite{schirmer2022non}.
In these methods, the activation sequences are often the result of CPSs entering successive states.
Although this problem appears very close to the one at hand, many techniques in this community rely on dedicated features, active and reactive power or peak current to name but a few, and problem-specific methods such as change or event detection \cite{schirmer2022non, faustine2017survey}.
Popular techniques include most notably hidden Markov models (HMM) \cite{lange2018variational}, where the states of a Markov model are not observed directly but are implicitly defined by a probability density function (pdf).
Deep learning architectures have also proven successful either using denoising autoencoders in a sequence-to-sequence fashion \cite{kelly2015neural} or convolutional neural networks \cite{zhang2018sequence}.
Overall, energy disaggregation rather focuses on pattern recognition such as device-specific power distribution or state transitions, putting the emphasis on the process and the machines instead of the underlying actuators.
This can be problematic in presence of flexible processes or event-based control, where the process changes and the actuator's activation sequences are not well separated.

A more general approach to this problem is through blind source separation (BSS), the action of retrieving a set of $S$ source signals from $M$ mixed signals.
BSS has received great attention over the years in multiple domains, from audio source separation \cite{hennequin2010nmf, wang2023probability}, energy disaggregation \cite{matsumoto2016energy} to fault detection and diagnosis \cite{wodecki2017local, liang2022impulsive, gabor2023non}.
The BSS problem is underdetermined if $M < S$.
This problem is often dealt with as a matrix factorization which consists in reconstructing data as the composition of two matrices. 
Two mathematical formulations stand out.
On the one hand, clustering can be used to separate operating phases in which sources are mixed, the two matrices then correspond to a \textit{mixing matrix} and \textit{source signals} different in each phase \cite{wang2021novel, loesch2008source, xie2019underdetermined}.
On the other hand, data can be reconstructed as a sparse \textit{representation} on a \textit{dictionary} \cite{matsumoto2016energy, mairal2009online}.

Early Line Spectra Estimators (LSE), inspired by Prony's method in the 1980s, exploit the fact that amplitude and phase estimation becomes least squares solvable when the frequencies are known \cite{stoica2000amplitude}.
This led to the popular subspace methods decomposing discrete data into signal and noise subspaces, among which figure the MUSIC \cite{schmidt1986multiple} and ESPRIT \cite{roy1989esprit} algorithms, and variants thereof either on-grid \cite{liu2022reconstruction, xu2022focusing, kiser2023real}, or off-grid with a parameterized sparse Fourier representation \cite{lasserre2015bayesian}.
Statistical decomposition can also be achieved through underdetermined independent component analysis (ICA) \cite{kim2004underdetermined, zheng2022identifiability}, although this technique is limited to sub-Gaussian signals and does not perform well in presence of discrete events.
Alternatively, sparse component analysis (SCA) \cite{gribonval2006survey, xu2019enhanced} first applies a sparsifying transform on data such as the short time Fourier transform (STFT) or wavelet transform (WT), the rationale being that sources are easier to separate in a lifted space where they exhibit noticeable differences.
Spectral decomposition techniques can also construct an orthonormal eigenbasis, onto which the projection of the data results in separated sources.
These techniques include singular value decomposition (SVD) \cite{zhao2009similarity, feng2013recent}, difference mode decomposition \cite{hou2023difference} as well as dynamic mode decomposition (DMD) \cite{schmid2010dynamic, proctor2016dynamic}.
In the time-frequency (TF) domain, Nonnegative Matrix Factorization (NMF) and Nonnegative Tensor Factorization (NTF) \cite{gabor2023non} are a natural choice when working on positive features such as spectrograms or bi-frequency maps \cite{wodecki2019impulsive}.
Although effective in isolating sequential dynamic behaviors, subspace methods fall short when the underlying signals do not have orthogonal spectra, as the embedding no longer represents the true sources, but their common characteristics.
Moreover, these techniques lack a mechanism to force the representation to lie in a binary space as expected in a multi-label classification problem.

To remedy this limitation, semi-binary NMF forces the representation to be binary \cite{matsumoto2016energy}.
Setting aside partially supervised implementations, popular in energy disaggregation \cite{matsumoto2016energy}, binarity can be enforced through means of regularization \cite{wodecki2017local} or directly using a coupled factorization method and relaxed alternating least squares (ALS) \cite{sorensen2022overlapping}.
The nonnegative requirement imposes the use of nonlinear transforms and hence phase removal.

An alternative consists in learning a shift-invariant dictionary based on the convolution operator \cite{zhou2016detection}.
Convolutional sparse coding strategies learn a sparse representation as its convolution with temporal patterns.
Traditional techniques include the shift-invariant sparse coding model from Grosse et al. \cite{grosse2012shift} and variants \cite{jas2017learning}. These models either use prior knowledge on the source signals (pre-computed dictionary with source responses), or learn the dictionary along with the representation \cite{dupre2018multivariate}.
Sub-dictionaries containing time shifted copies of the initial dictionary were recently proposed by Wang et al. \cite{wang2023novel}, although this discrete approximation fails to capture events lying off-grid.

At last, the mentioned techniques require the number of sources to be known.
For dimensionality reduction, this number can be approximated \cite{loesch2008source}, yet this often boils down to rank estimation \cite{antoni2011second}.
Noisy data may induce a large Pareto front though.
This makes rank estimation highly imprecise.
Another possible cause of error in rank estimation is the use of nonlinear transforms.
Proposals have been made in previous work to decompose signals with an accurate estimation of the number of sources, either using tracking for non-stationary cases \cite{delabeye2023sequential} or matrix factorization \cite{delabeye2022unsupervised}.

To the best of the authors' knowledge, to date there is no matrix factorization method in the literature capable of recovering the exact activation sequences from a mixed signal under the constraints considered in this paper:
fully unsupervised underdetermined blind source separation with unknown number of sources, applied to additive, potentially correlated, periodic, ergodic and quasi-stationary source signals with arbitrary noise distribution and no prior knowledge, resulting in a complex dictionary with a binary representation.
Our main contribution is twofold, (i) we propose a novel formulation circumventing the pitfalls arising in semi-binary matrix factorization in a complex vector space, and (ii) a greedy algorithm to learn both the dictionary and the sparse representation.

The remainder of this paper is set out as follows.
After formally describing the particular underdetermined blind source separation problem this paper is concerned with in \prettyref{sec:pbformulation}, a clustering-based two-step algorithm is proposed.
In \prettyref{sec:centroids}, we show that the exact recovery of source activation sequences from non-intrusive sensor data is intrinsically tied to the presence of problematic phase shifts, the causes of which are detailed.
A solution to the complex semi-binary matrix decomposition problem is found in \prettyref{sec:csbmf} by carefully resynchronizing the sources.
This method is finally verified on synthetic data and compared to existing methods in \prettyref{sec:results_num}. Experimental validation is undertaken in \prettyref{sec:results_exp}, in which we introduce the CAFFEINE dataset \cite{delabeye_romain_2023_8351431}.
Limitations and perspectives are discussed in \prettyref{sec:discussion}, before concluding.

\section{Methods}
\label{sec:methods}

\subsection{Problem formulation}
\label{sec:pbformulation}

A production process is a sequence of \textit{operations}. 
Each of these involves a collection of \textit{actuators}. 
Operations are therefore successive and cannot overlap.
An actuator comprehends all the elements of a connected power chain, actionable simultaneously and controlled as a whole.
These actuators produce \textit{source} signals when aggregated by a sensor, resulting in a time series of sequential mixed signals.
An actuator is a source $s$, switched on ($1$) and off ($0$) according to its activation sequence $\mathrm{ACT}_{s}(t)$ over time.
Let $\mathcal{C}$ denote the alphabet of all distinct operations in data, each operation involving simultaneous sources.
An operation is a group of sources $c \in \mathcal{C}$, later denoted \textit{cluster}.
It is attributed an activation status $\mathrm{OPS}_{c}(t)$.

The short time Fourier transform (STFT) is used in this paper, as it is well suited to the study of piecewise stationary signals.
Its definition is recalled for a signal $x$ of finite support, uniformly sampled over time with frequency $f_s$.
The STFT can be viewed as a sliding discrete Fourier transform (DFT) applied to partially overlapping windows with hop size $H$, using an analysis window $\mathrm{w}$ with size $W$, indexed by discrete time step $m$, and frequency bin index $k$ associated with discretized pulse $\bm{\omega}$, with $\forall k \in \llbracket 0, W-1 \rrbracket, \omega_k = \frac{2 \pi k}{W}$:

\begin{equation}
    STFT\{x\}[m, k] = \sum_{n=0}^{W-1} x[n] \mathrm{w}[n - m] e^{-\jmath \frac{2 \pi k}{W} n}
\end{equation}

The time-shift theorem of the DFT is used to shift the signal in time while remaining in the frequency domain.
The time-shift operator $\bm{S}_{\Delta}$ for a time difference $\Delta$ is defined as:

\begin{equation}\label{eq:time_shift_operator}
  \bm{S}_{\Delta} = \diag \Big ( \big ( e^{- \jmath \frac{2 \pi k}{W} \Delta} \big )_{0 \leq k < W} \Big )
\end{equation}

Let $\bm{X} \in \mathbb{R}^{T^\prime}$ denote a univariate time series representing $T^\prime$ sensor measurements sampled at frequency $f_s$.
$\bm{X}$ is produced by $S$ actuators sequentially activated, in use in $N_{ops}$ distinct operations, with $N_{ops} \geq S$.
From time series $\bm{X}$, the feature matrix $\bm{Z} \in \mathbb{C}^{W \times T}$ is computed using the STFT, where $W$ is the number of frequency bins (and window size) and $T$ is the number of feature samples (time windows) in the TF domain, with $T' \geq T$.

The retrieval of the sources' descriptors and activation sequences can be sought as the optimal solution to an underdetermined semi-binary matrix decomposition problem \cite{wodecki2017local}. Here, $S$ sources are mixed over a single channel.
Undetermined blind source separation is an inverse problem, ill-posed in that the matrix factorization (regardless the formulation) does not admit a unique stable solution.
This issue is often overcome through regularization and constraints on the dictionary, the representation or both.
A classic formulation is as a sparse dictionary learning problem \cite{wodecki2017local, liang2022impulsive, wodecki2019impulsive, wang2023novel}:

\begin{equation}
    \argmin_{\substack{
        \bm{D} \: \in \: \mathcal{D}, \: \bm{R}
        \\
        \bm{R} \geq 0
        }
    } 
    \sum_{m=1}^T 
    \Bigg (
    \Psi(\bm{Z}_m - \bm{D} \bm{R}_m)
    + \mathcal{R}_{sparse}(\bm{R}_m)
    + \mathcal{R}_{binary}(\bm{R}_m)
    \Bigg )
\end{equation}

where $\bm{D} \in \mathbb{R}^{W \times N}$ and $\bm{R} \in \mathbb{R}^{N \times T}$ are the \textit{dictionary} and \textit{representation} (multi-labels over time) to be learnt.
$\mathcal{R}_{sparse}$ is a sparsity-promoting penalty, typically the Least Absolute Shrinkage and Selection Operator (LASSO). 
Binary solutions are enforced either using a regularizer $\mathcal{R}_{binary}$ \cite{darabi2018bnn+} or an alternative to the least-squares functional \cite{wodecki2017local}.
$\mathcal{D}$ is a set of constraints on $\bm{D}$, necessary to prevent the penalties on $\bm{R}$ from being compensated by larger elements in $\bm{D}$ due to the coupled functional, as both are jointly optimized.
Labels are constrained to positive values for convenience.
Indeed, allowing signed labels would result in implicitly defined actuators, i.e., as the presence of a collection of actuators (positive labels) while excluding some others (negative labels).

The signals produced by the sources are assumed to be quasi-stationary and ergodic.
Stationarity is a strong assumption, here its weak form is preferred, in which only the mean and the covariance of the process must be time-invariant and finite \cite{priestley1988non}.
In other words, for each operation, the steady state lasts long enough for the transient response to have a negligible impact on the mean and variance of its descriptor.

Input signal is thus piecewise consistent and there exists a \textit{dictionary} \mbox{$\bm{C} \in \mathbb{C}^{W \times N}$} of \textit{atoms} $[ \bm{C}_1, ..., \bm{C}_{N} ]$, also called \textit{centroids} or \textit{descriptors}, properly describing the stationary state of each operation.
In a complex vector space, phase shifts occur and a single operation may be present in more than one configuration, i.e., the sources' time shifts may differ from a realization to another.
Hence $N \geq N_{ops}$.
A subset $\bm{\Tilde{C}} \in \mathbb{C}^{W \times S}$ of these centroids describes the actuators isolated from all others.
It is here assumed that each actuator is seen at least once alone in the dataset.

Similarly, the operation the underlying system is in at each time step is represented by the one-hot encoded label matrix $\bm{\breve{L}} \in \{0, 1\}^{N_{ops} \times T}$, or $\bm{L} \in \{0, 1\}^{N \times T}$ to distinguish all configurations. Let $\lVert . \rVert_0$ denote the $\ell_0$ norm, $\forall m \in \llbracket 0, T-1 \rrbracket$, $\lVert \bm{L}_m \rVert_0 = 1$.
Whereas the activation statuses of a system's actuators over time are gathered in a multi-hot encoded matrix $\bm{\Tilde{L}} \in \{0, 1\}^{S \times T}$, with $\lVert \bm{\Tilde{L}}_m \rVert_0 \geq 1$.
Thus the time series $\bm{L}^c$ and $\bm{\Tilde{L}}^s$ indicate whether the system is in operation $c$ and uses actuator $s$ at each time step. 
Matrix superscripts and subscripts represent vectors lying in the row and column spaces respectively.

Here we lay out the assumptions, humorously coined the ten plagues of unsupervised complex semi-binary matrix factorization, which arise from the blind source separation problem and from the application, recovering actuator activation statuses in an industrial environment from non-intrusive sensors and without supervision.

\noindent\begin{minipage}[t]{.5\linewidth}
\begin{assumption}\label{P:1}
Fully unsupervised
\end{assumption}
\begin{assumption}\label{P:2}
Unknown number of sources
\end{assumption}
\begin{assumption}\label{P:3}
Underdetermined ($S < N$)
\end{assumption}
\begin{assumption}\label{P:4}
Periodic source signals
\end{assumption}
\end{minipage}%
\begin{minipage}[t]{.5\linewidth}
\begin{assumption}\label{P:5}
Potentially correlated sources
\end{assumption}
\begin{assumption}\label{P:6}
Quasi-stationary source signals
\end{assumption}
\begin{assumption}\label{P:7}
No prior knowledge on sources
\end{assumption}
\begin{assumption}\label{P:8}
Any noise distribution
\end{assumption}
\end{minipage}

\vspace{\dp0}

\begin{assumption}\label{P:9}
Complex dictionary and binary representation
\end{assumption}

\begin{assumption}\label{P:10}
Each sensor measures an extensive property
\end{assumption}

The above-described problem uses a dictionary, each atom involving a group of sources.
Due to binarity, the decomposition result only makes sense if each atom in the minimal set $\bm{\Tilde{C}}$ represents exactly one source tied to an actuator.
In the method proposed in this paper, this property is ensured using time series clustering in conjunction with \prettyref{P:11}.

\begin{assumption}\label{P:11}
Each source appears alone at least once in data
\end{assumption}

\begin{figure}[H]
    \centering
    \includegraphics[width=1.\linewidth]{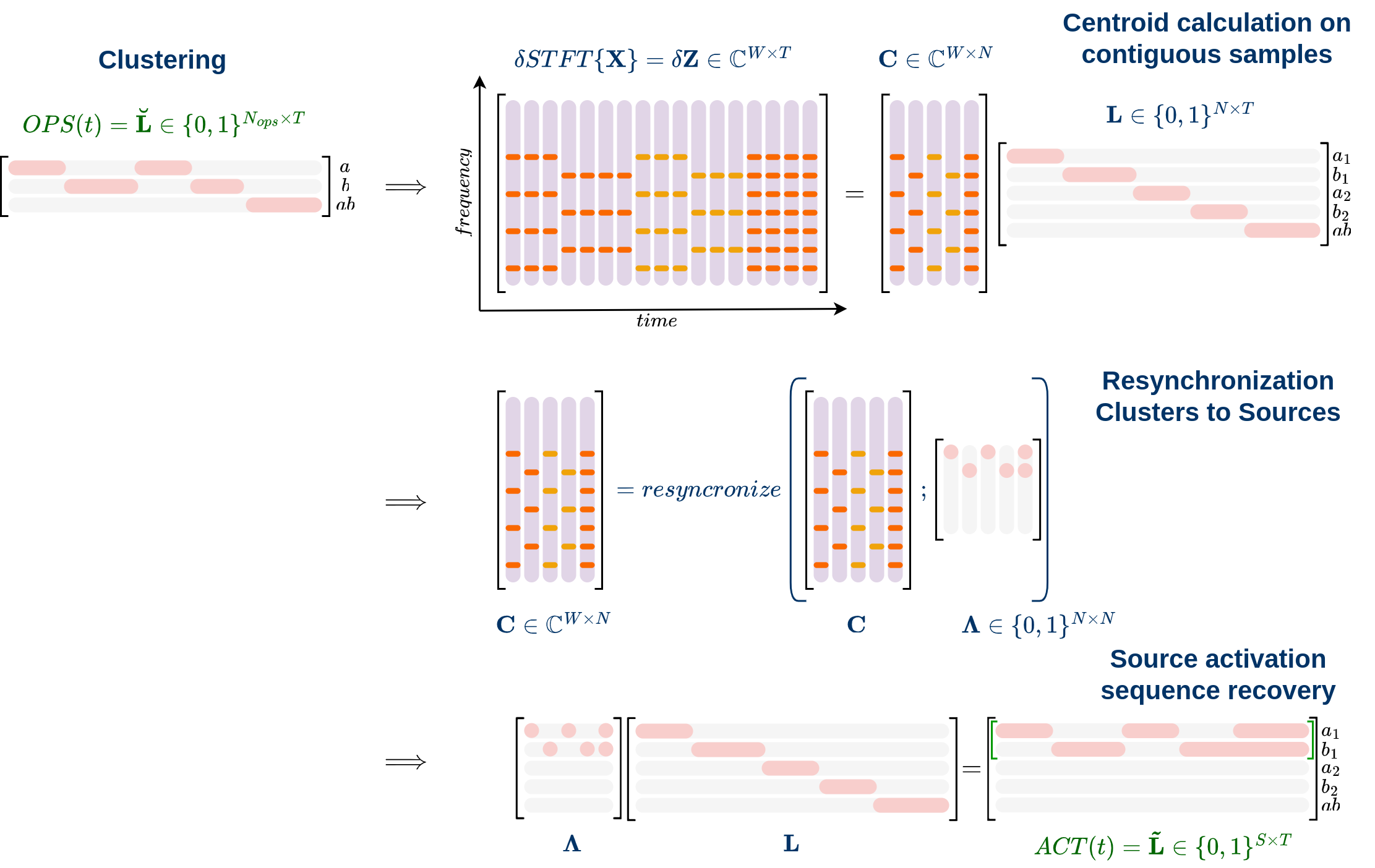}
    \caption{Source activation recovery steps}
    \label{fig:notations_wrapup}
\end{figure}

The proposed method and associated matrix notations are summarized in \prettyref{fig:notations_wrapup}.

\subsection{Clustering data into successive operations}
\label{sec:clustering}

A first step of the proposed method consists in segmenting the signal into successive operations. Clustering techniques prove useful in addressing this problem, by breaking down data into groups of similar objects \cite{rai2010survey}.
In this context, clustering is applied to multivariate time series, and boils down to grouping timestamps according to a similarity measure between sample vectors. 
In hard clustering, each data point belongs to a single partition, whereas soft clustering allows for partial membership to different clusters.
Time series clustering has been extensively tackled in the literature, and widely applied to structural health monitoring in particular, from fault detection \cite{gangsar2020signal, rai2017bearing} to dynamic load identification \cite{liu2022distributed}.

Despite its simplicity, the \textit{k-means} algorithm is considered here as a suitable candidate to segment the data when the number of operations $N$ is known. 
Otherwise $N$ is estimated from a dendrogram or by maximizing a criterion.
Alternatives in which the number of clusters is not required exist in the literature, yet it is often replaced by other hyperparameters.
For instance, clustering techniques such as \textit{DBSCAN} or its variant \textit{OPTICS} \cite{wang2021novel} may be better suited in this setting.

In the proposed approach, clusters are sought using any relevant feature, obtained through linear or nonlinear transforms, so long as it exhibits piecewise stationarity.
This paper focuses on periodic signals, for which phase-invariant spectral descriptors are well suited to clustering tasks (e.g., spectrogram, STFT magnitude, first four statistical moments applied to sliding windows, or any alternative suitable for stationary periodic signals).

Eventually, the one-hot encoded operation labels $\bm{\Breve{L}}$ are obtained.
The centroids associated with these labels, cluster-wise average features, are of little use though.
Indeed, nonlinear transformations were introduced, making the centroids lose their additive property (\prettyref{P:11}).
That is, the centroid of a superimposed state must be equal to the sum of the centroids of its states.
An operator is hence introduced in \prettyref{sec:centroids} to deduce consistent centroids suitable for the decomposition part.
\subsection{Computing consistent complex centroids despite phase shifts}
\label{sec:centroids}

In this section, centroids are computed with a view to later expressing each operation as the sum of other operations.
We seek a transform (applied to signal $\bm{X}$) such that (i) stationarity assumption \prettyref{P:6} is preserved in the feature space, and (ii) the additive property of the signal (\prettyref{P:10}) is kept throughout the transform.
To that end, we propose a modified time-shifted STFT operator, denoted $\delta STFT\{.\}[m, k]$.
The proposed linear operator leaves the phase unchanged across windows operating on the same signal.

\begin{figure}[h!]
    \centering
    \includegraphics[width=.5\linewidth]{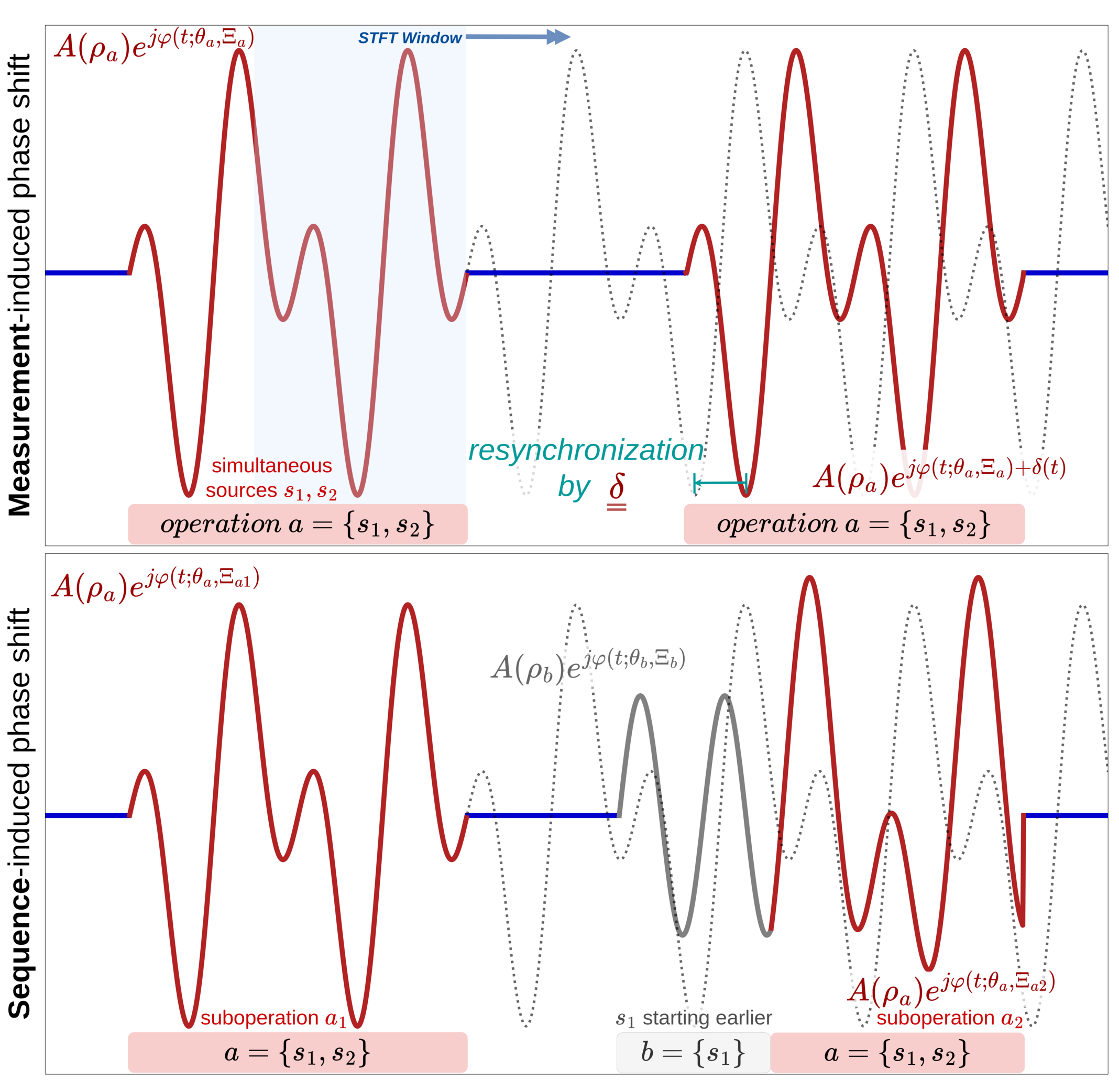}
    \caption{Phase discrepancies translated in the time domain}
    \label{fig:phasePb_time}
\end{figure}

Indeed, when expressing the input signal as a sum of sources, we notice that phase shifts occur from a window to another.
Let $x$ be a stationary signal over a single operation containing $S$ simultaneously active harmonic sources. Each source $s$ produces a signal with amplitude $A_s$, pulse $\omega_s$, reference phase $\varphi_s$ and phase shift $\xi_s$ with respect to this reference.
From this definition, an operation is characterized by source-dependent invariant parameters $\theta = \{[\omega_s, \varphi_s]\}_{s=1}^S$ and $\rho = [A_1, ..., A_S]^T$.
Because an operation starts as the consequence of an event, the activation or deactivation of a source, entering an operation resets one of the time shifts $\xi_s$.
This means that parameters $\Xi = [\xi_1, ..., \xi_S]^T$ are invariant only throughout one sub-operation, i.e., one realization of this operation.
Parameters $\rho, \theta, \Xi$ are unknown.
As illustrated in \prettyref{fig:phasePb_time}, every time a source is switched off then on again after a non-integer number of periods, the initial phase of the source signal rotates.
As a result, the $k$-th component of the DFT of $x$ over a window of the STFT, starting at time step $m_0$ and ending at time step $m_1$, indexed by $m \in \llbracket m_0, m_1 \rrbracket$ has the form:

\begin{equation}
\begin{split}
    DFT_k\{x_{mH:mH+W-1; \rho, \theta, \Xi}\} 
  &= \sum_{s=1}^{S} \sum_{n=0}^{W-1} 
  A_s e^{\jmath \big ( \omega_s n + \varphi_s + \xi_s - \omega_k n - \omega_k mH \big )}
  \\
  &= A(\rho) e^{\jmath \bm{\varphi}(m; \theta, \Xi)}
\end{split}
\end{equation}

In order to recover one reference DFT for a sub-operation, we estimate the time lag $\delta$ by which the measured DFT should be shifted so the resulting feature remains constant over $\llbracket m_0, m_1 \rrbracket$.
Indeed, the event having led to this sub-operation may have occurred after a non-integer number of hops.

\begin{equation}\label{eq:DFT0}
\begin{aligned}
    &
    \frac{\partial}{\partial m} DFT\{ x_{\delta + m H : \delta + m H + W - 1}; \rho, \theta, \Xi \} 
    &= 0
    \\[5pt]
    \iff 
    &
    \frac{\partial}{\partial m} \bm{S}_{\delta + m H} DFT\{ x_{0:W-1}; \rho, \theta, \Xi \} 
    &= 0
    \\[5pt]
    \iff 
    &
    - (H + \frac{\partial \delta}{\partial m}) diag(\bm{\omega}) \bm{S}_{\delta + m H} DFT\{ x_{0:W-1}; \rho, \theta, \Xi \} 
    &= 0
\end{aligned}
\end{equation}

Then either $A(\rho) = 0$ and shifting is irrelevant (trivial solution), $\delta$ is time linear, or the phase cancels out. Solving for $\delta$ yields its least squares estimate:

\begin{equation}\label{eq:delta_stftshift_generalvalue}
    \delta = \frac{\bm{\omega}^T}{\bm{\omega}^T \bm{\omega}} \big ( \bm{\varphi}(0; \theta, \Xi) + ( 2 \pi p - m H ) \mathbbm{1}_{W}^T \big )
\end{equation}

where $p \in \mathbb{Z}$ and $\mathbbm{1}_W^T$ is the one-vector of size $W$. As expected, $\delta$ depends in turn on $\theta$ and $\Xi$, and $\delta[m] = \delta_0(\theta, \Xi) - m H$.
The phase only conveys noise in the low energy regions of the spectrum though.
Since the STFT is naturally sparse, the least squares solution is a very poor estimator in this case.
Instead, we simply measure the phase $\varphi(m; \theta, \Xi)[K]$ at the maximum of amplitude --- excluding the DC component --- to estimate $\hat{\delta}[m]$:

\begin{equation}\label{eq:delta_stftshift_maxamp}
    \hat{\delta}[m] = \frac{\bm{\varphi}(m; \theta, \Xi)[K]}{\omega_K}
\end{equation}

Alternatively, $\hat{\delta}[m]$ can be estimated using \prettyref{eq:delta_stftshift_generalvalue} on the dominant frequencies.
Or using an optimal state estimator on the successive realizations of $\delta[m]$ in the sub-operation $\llbracket m_0, m_1 \rrbracket$, by noticing that $\delta[m]$ has strictly the same linear dynamics (provided that the phase is carefully unwrapped) and statistical properties as $\bm{\varphi}(m; \theta, \Xi)$.

A collection of centroids $\bm{C} = [\bm{C}_1, ..., \bm{C}_N]$ is obtained by averaging out the samples in every sub-operation with contiguous samples, with $N \geq N_{ops}$.
Each sub-operation $j$ holds sequence- and signal-dependent phase differences $\Xi^{(j)}$ with respect to the operation's reference $\theta$.
These different configurations for a single operation cannot be easily untangled, as illustrated on \prettyref{fig:phasePb_time}.
Decomposition is thus run on these centroids, and the optimization process proposed in Section \ref{sec:csbmf} will naturally recombine these sub-operations using source resynchronization.

Spectral leakage constitutes yet another cause of error in estimating an operation's reference DFT, often due to a non-integer number of periods present in a window.
Energy ends up distributed across the spectrum which results in undesired frequency components.
Choosing an appropriate window function $\mathrm{w}$ alleviates this phenomenon.

A procedure to compute the $\delta$STFT is proposed in Algorithm \ref{alg:deltastft}.

\begin{algorithm}[H]
\caption{proposed $\delta STFT\{.\}[m, k]$}\label{alg:deltastft}
\KwIn{
Time series $\bm{X} \in \mathbb{R}^{T^\prime}$; clustering labels $\bm{\Breve{L}} \in \{0, 1\}^{N_{ops} \times T}$\;
}
\KwResult{
Piecewise constant DFTs over time $\delta \bm{Z} \in \mathbb{C}^{W \times T}$, and centroids $\bm{C} \in \mathbb{C}^{W \times N}$\;
}

\textbf{Step 1: } Apply the STFT as $\bm{Z} \in \mathbb{C}^{W \times T}$, $\bm{Z} = STFT\{\bm{X}\}$ \;

\textbf{Step 2: } Extract features for decomposition

\For{$0 \leq m < T$}
    {
    Measure the phase at the maximum of magnitude 
    
    (excluding the DC component)   

    $\varphi_{max} = \angle \displaystyle\argmax_{
    \bm{z} \in \{\forall k > 0,\: \bm{Z}_m^k\}
    } {\lvert \bm{z} \rvert}$\;

    corresponding to the frequency bin with pulse $\omega_{max}$

    and estimated time shift $\hat{\delta} = \frac{\varphi_{max}}{\omega_{max}} \frac{f_s}{2}$
    \;
    
    Time shift the phase accordingly $\delta\bm{Z}_m \gets \bm{S}_{\hat{\delta}} \bm{Z}_m$\;
    }

    Lift clustering labels as $\bm{L}$ to represent only contiguous samples\;

\textbf{Step 3: } Compute sub-centroids as $\bm{C} = \delta\bm{Z} \bm{L}^T diag \Big ( \big (\frac{1}{\lVert \bm{L}^g \rVert_0} \big )_{1 \leq g \leq N} \Big )$\;

\end{algorithm}

Lastly, if present, the vector with the least root mean square (RMS) is removed from $\bm{C}$ as it relates to the \textit{stand-by} operation. 
This operation corresponds to background noise or a persistent component detrimental to the decomposition problem (much like the neutral element of a set).
Decomposing centroids instead of samples greatly reduces the computational complexity, as the matrix factorization no longer depends on the number of samples but the number of operations.
The use of centroids is also more robust to noise.

\subsection{Matrix decomposition as a resynchronization problem}
\label{sec:csbmf}

In this section, the goal is to retrieve the actuators' activation sequences $\bm{\Tilde{L}} \in \{0, 1\}^{S \times T}$, given the elicited centroids $\bm{C} \in \mathbb{C}^{W \times N}$.

In this paper, we propose a convenient parameterization for both the dictionary and the representation, in which the optimization problem can be effectively regularized.
Indeed, by computing the dictionary as a set of centroids from the $\delta STFT$, the atoms are forced to retain physical properties.
This dictionary is then parameterized in the time lags $\bm{\Delta} \in \mathcal{I}_{\bm{\Delta}}$ required to optimally reconstruct each operation's centroid.
Since the sources are periodic, so is the process of resynchronizing each source in a sum.
The optimization could hence be carried out on $\cup_{c=1}^N [-\frac{\hat{T}^{(c)}}{2}, \frac{\hat{T}^{(c)}}{2}[^{N}$, where $\frac{1}{\hat{T}^{(c)}}$ is the estimated fundamental frequency of each atom.
Similarly, atoms are expressed as linear combinations of others.
That is, the content of each operation is stacked in column form in matrix $\bm{\Lambda} \in \mathcal{I}_{\bm{\Lambda}}$, where each column is a collection of operations meant to be learnt in place of the sources' activation sequences.
$\bm{\Lambda}$ thus constitutes a Rosetta Stone, translating each operation as its content in terms of other operations, or directly in terms of the underlying sources when the maximal decomposition is reached.
The sought solutions lie in $\{0, 1\}^{N \times N}$.

Even under these conditions, decomposition remains challenging.
In particular, source resynchronization is well known to be highly non-convex and entails a combinatorial number of spurious minimizers \cite{gossard2022spurious}.
This phenomenon is clearly illustrated on a synthetic use case in \prettyref{fig:residualmatrix}.
Moreover, the dictionary is redundant down to the time shifts, hence there exists a myriad of global minimizers for the representation as well \cite{wang2023novel}, albeit only a handful are relevant.

We thus propose a novel formulation to overcome the outlined difficulties.
Constraints are lifted to begin with.
Tikhonov regularization is applied to the time shifts, allowing for an unconstrained optimization on $\mathcal{I}_{\bm{\Delta}} = \mathbb{R}^{N \times N}$ directly.
This does not prevent the existence of a combinatorial number of local minima though.
The representation suffers from many more causes of indetermination.
Using the regularization approaches in \prettyref{eq:main}, detailed in \ref{apx:opt}, the constraint on the representation can be lifted to optimize on $\mathcal{I}_{\bm{\Lambda}} = \mathbb{R}^{N \times N}$, instead of $\{0, 1\}^{N \times N}$ which is NP-hard.

In the frequency domain, if an operation $c$ with descriptor $\bm{C}_c$ can be decomposed as a sum of operations $\bm{\Lambda}_c$ given time-shifts $\bm{\Delta}_c$, then the $\mathbb{C}$SBMF is formulated as an optimization problem:

\begin{equation}
\label{eq:main}
    \begin{split}
    \inf_{\substack{
        \bm{\Delta} \in \mathcal{I}_{\bm{\Delta}}, 
        \\
        \bm{\Lambda} \in \mathcal{I}_{\bm{\Lambda}},
        }
    }
    \:
    &
    \sum_{c=1}^{N} \Bigg (
    \Big \lVert \bm{C}_c - \sum_{i=1}^N \bm{S}_{\Delta_c^i}\bm{C}_i \Lambda_c^i \Big \rVert_2^2
    \\
    &
    + \lambda \mathcal{L}_{col}(\bm{\Lambda}_c)
    + \mathcal{E} \mathcal{T}(\bm{\Lambda}_c)
    + \beta \mathcal{B}_2 \big ( \bm{\Lambda}_c \big )
    + \Gamma \lVert \bm{\Delta}_c \rVert_2^2
    \Bigg )
    + L \mathcal{L}_{row}(\bm{\Lambda})
    \end{split}
\end{equation}

\noindent\begin{minipage}[c]{.4\linewidth}%
\begin{equation}
  \mathcal{L}_{col}(\bm{\Lambda}_c) = \big \lVert \bm{\Lambda}_c \big \rVert_{p}
\end{equation}
\end{minipage}%
\begin{minipage}[c]{.6\linewidth}%
\begin{equation}\label{eq:def_regularizer_triangle}
  \mathcal{T}(\bm{\Lambda}_c) = \frac{\big \lVert \bm{\Lambda}_c \big \rVert_p}{\lVert \bm{C}_c \rVert_2^2}
\end{equation}
\end{minipage}
\vspace{\dp0}

\begin{minipage}[c]{.38\linewidth}%
\begin{equation}
  \mathcal{L}_{row}(\bm{\Lambda}) = \big \lVert \bm{\Lambda}^T \big \rVert_{2, p}
\end{equation}
\end{minipage}
\begin{minipage}[c]{.55\linewidth}%
\begin{equation}
  \mathcal{B}_2(\bm{\Lambda}_c) = \big \lVert \frac{1}{2} \mathbbm{1}  - \lvert \bm{\Lambda}_c - \frac{1}{2} \mathbbm{1} \rvert \big \rVert_2
\end{equation}
\end{minipage}
\vspace{\dp0}
\bigskip

where $\mathcal{L}_{col}$, $\mathcal{T}$, $\mathcal{B}_2$, $\lVert \bm{\Delta}_c \rVert_2^2$ and $\mathcal{L}_{row}$ denote the penalties associated with the regularization coefficients $\lambda$, $\mathcal{E}$, $\beta$, $\Gamma$, and $L$ respectively. 
The least squares functional is denoted $F(\bm{\Delta}, \bm{\Lambda})$.
$\lVert . \rVert_p$ is the $\ell_p$ norm (for $0 < p \leq 1$), and $\lVert \bm{\Lambda} \rVert_{2, p} = \Big ( \sum_{i=1}^N \lVert \bm{\Lambda}_i \rVert_2^p \Big )^{\frac{1}{p}}$ is $\ell_{2, p}$ matrix norm.
$\mathbbm{1}$ is the one vector.

Sparsity is promoted in two ways, column-wise with the $\ell_p$ norm to sparsely decompose each atom, and row-wise with an $\ell_{2, p}$ penalization to fight the dictionary's redundancy.
The latter is motivated by the fact that the $\ell_{2, p}$ norm is an adequate approximation of the $\ell_{2, 0}$ norm which is the exact number of non-empty rows \cite{zhao2018trace}.
Here, the number of nonzero rows in $\bm{\Lambda}$ is the estimated number of sources.

A particularity of the proposed method is that the dictionary was built based on clustering.
In compressed sensing, this is the worst choice for a dictionary since the most sparse solution is actually the one involving all atoms.
That is, the trivial solution $\bm{\Lambda} = \bm{I}$ (identity) corresponds exactly to the clustering result.
For instance, $abc=ab+c$ is more sparse than $abc=a+b+c$, yet the latter is sought.
The competing objectives $\mathcal{L}_{col}$ and $\mathcal{L}_{row}$ remedy this situation.
This calls for a subtle choice for $L$ though, making $\mathcal{L}_{row}$ always greater than $\mathcal{L}_{col}$, and thus prioritizing the estimation of the number of sources.

Another unorthodox regularization term $\mathcal{T}$ is proposed.
This term endows the column-wise sparse regularization parameter with a bias decreasing as the squared $\ell_2$ norm of a suspected source increases.
This penalty is crucial in that it avoids a pitfall arising in complex vector spaces: phase reversal.
Indeed, phase resynchronization induces rotations.
Hence in a resynchronized sum, vectors can flip and cancel out other components. As an example, \mbox{$ab=a+b$} could be strictly equivalent (in cardinality and residual on $F(\bm{\Delta}, \bm{\Lambda})$) to \mbox{$a=ab+b$}.
An arbitrary rule is required to distinguish these minima, since these combinations are algebraically equivalent (all satisfy the triangle inequality).
Here, $\mathcal{T}$ is designed so the energy of the sum is higher than that of any of its constituents.
The physical interpretation of $\mathcal{T}$ corresponds to the assumption that a collection of systems operated concurrently cannot draw less power than any of the underlying systems operated alone.
While unlikely, independent sources may damp each other out and lead to a less energetic sum violating this assumption, as sometimes occurs in vibration mechanics.
If this phenomenon is identified, moving $\lVert \bm{C}_c \rVert_2^2$ to the numerator of $\mathcal{T}$ reverses the order.

A binarity penalty $\mathcal{B}_2$ is added, similarly to the one proposed by Darabi et al. \cite{darabi2018bnn+}.
Sparse regularization leads to a suitable approximation of $\bm{\Lambda}$, up to a factor since the $\ell_p$ norm draws the minimum towards zero.
Binary regularization rectifies this, as well as any noise-originated discrepancy in the estimation of $\bm{\Lambda}$. 

Overall, given a base dictionary $\bm{C} \in \mathbb{C}^{W \times N}$ and optimal regularization coefficients, \prettyref{eq:main} admits non-equivalent minima on $\mathbb{R}^{N \times N} \times \mathbb{R}^{N \times N}$.
The optimal representation $\bm{\Lambda}$ is meaningful in that it corresponds to the maximal binary decomposition of each atom.
This claim is supported by the generic example presented in \ref{apx:opt}, where the regularization mechanisms and their effect on the minima's locations are detailed.
Source activation sequences are finally recovered as $\bm{\Tilde{L}} = \bm{\Lambda} \bm{L}$.


\subsection{Practical implementation}

Despite the regularization terms introduced in \prettyref{eq:main}, the cost function is still highly non-convex and entails spurious local minimizers.
There are also multiple hyperparameters on which depends the relevance and accuracy of the representation.
Hyperparameter optimization has been extensively studied in the literature \cite{fuentes2021equation}, yet the reliability of these methods remains limited, especially as the number of parameters to tune grows.

Alternate optimization strategies can aim towards the sought minimum \cite{yu2021acoustic}, separately and gradually optimizing for the time shifts $\bm{\Delta}$ and the representation $\bm{\Lambda}$.
Convergence cannot be guaranteed though.
The effectiveness of such methods is therefore limited, especially as the time shifts of a combination do not inform on those of another combination.

For these reasons, optimizing with respect to both the time lags and the combinations at once may be too big a leap to ensure convergence towards the desired optimum.
On an important note, any suboptimal solution to \prettyref{eq:main} is completely useless for classification, as composite operations could be assigned a label distinct from the sources they contain.

Industrial applications come with a silver lining though.
Sensors are often limited to the monitoring of a few systems at a time (which limits the number of sources), and these systems may not use their actuators in all possible configurations (which limits the number of operations).
We hence advocate for a greedy algorithm to optimize for $\bm{\Lambda}$. 
Indeed, the entire parameter space is known and can be discretely mapped in a tractable way so long as the number of operations is reasonable (application-specific).
That is, for each possible decomposition of operation $c$ into a group of operations $\mathcal{G}$ indexed by $g$ with $g_{(2)} = \bm{\Lambda}^{(g)}_c$ (notation for an integer expressed in base $2$), a residual $r_g^c$ is compared to a threshold $\tau$, an energy bound on additive noise, to accept or reject the decomposition.

Every element of the residual matrix $\bm{R} \in \mathbb{R}^{N \times 2^N}$ is found as the solution to the resynchronization problem between vectors $\bm{C}_c$ and $\{\bm{C}_i \}_{i \in \mathcal{G}}$:

\begin{equation}
    r_g^c = 
    \displaystyle\inf_{\{\Delta_c^i\}_{i \in \mathcal{G}} }
    \Big \lVert \bm{C}_c - \sum_{i \in \mathcal{G}} \bm{S}_{\Delta_c^i} \bm{C}_i \Big \rVert _2^2
\end{equation}

\begin{figure}[H]
  \centering
  \includegraphics[width=1.\linewidth]{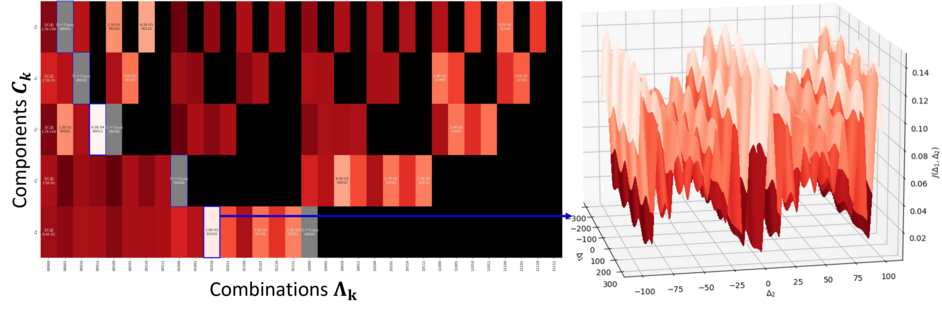}
  \caption{Residual matrix (left), after optimal source resynchronization (right)}
  \label{fig:residualmatrix}
\end{figure}

The Broyden–Fletcher–Goldfarb–Shanno (BFGS) algorithm \cite{fletcher2013practical} is used to compute these residuals.
As a result, the residual matrix $\bm{R}$ illustrated in Figure \ref{fig:residualmatrix} is obtained, each row indicating the possible combinations in $\{\bm{\Lambda}^{(g)}_c\}_{g = 0}^{2^N}$ for a given operation with centroid $\bm{C}_c$.

The parameter space results from the following heuristic.
For each operation $c \in \llbracket 1, N \rrbracket$ and combination $g \in \llbracket 1, 2^N-1 \rrbracket$ carrying indices $\mathcal{G}$:

\begin{itemize}
    \item If $g = 0$, then $r_g^c = \lVert \bm{C}_c \rVert_2^2$ is the squared norm.
    \item If $g_{(2)} = (2^k)_{(2)}$, then $r_g^c = 0$ (trivial decomposition).
    \item If $\lVert \bm{C}_c \rVert_2 > \sum_{i \in \mathcal{G}} \lVert \bm{C}_i \rVert_2$, then exclude $g$ (triangle inequality unsatisfied).
    \item If $\lVert \bm{C}_c \rVert_2^2 < \displaystyle\max_{i \in \mathcal{G}} \lVert \bm{C}_i \rVert_2^2$, then exclude $g$ (energy-based ordering).
\end{itemize}

In the absence of the sparsity regularizers, multiple admissible combinations may be found.
These minimizers bear different residuals due to noise and other sources of uncertainty, albeit centroids are naturally resilient in that respect.
For this reason, the decomposition returned by the proposed technique corresponds to the minimum number of sources to begin with, and only then the lowest residual is sought.

\begin{figure}[H]
    \centering
    \includegraphics[width=.8\linewidth]{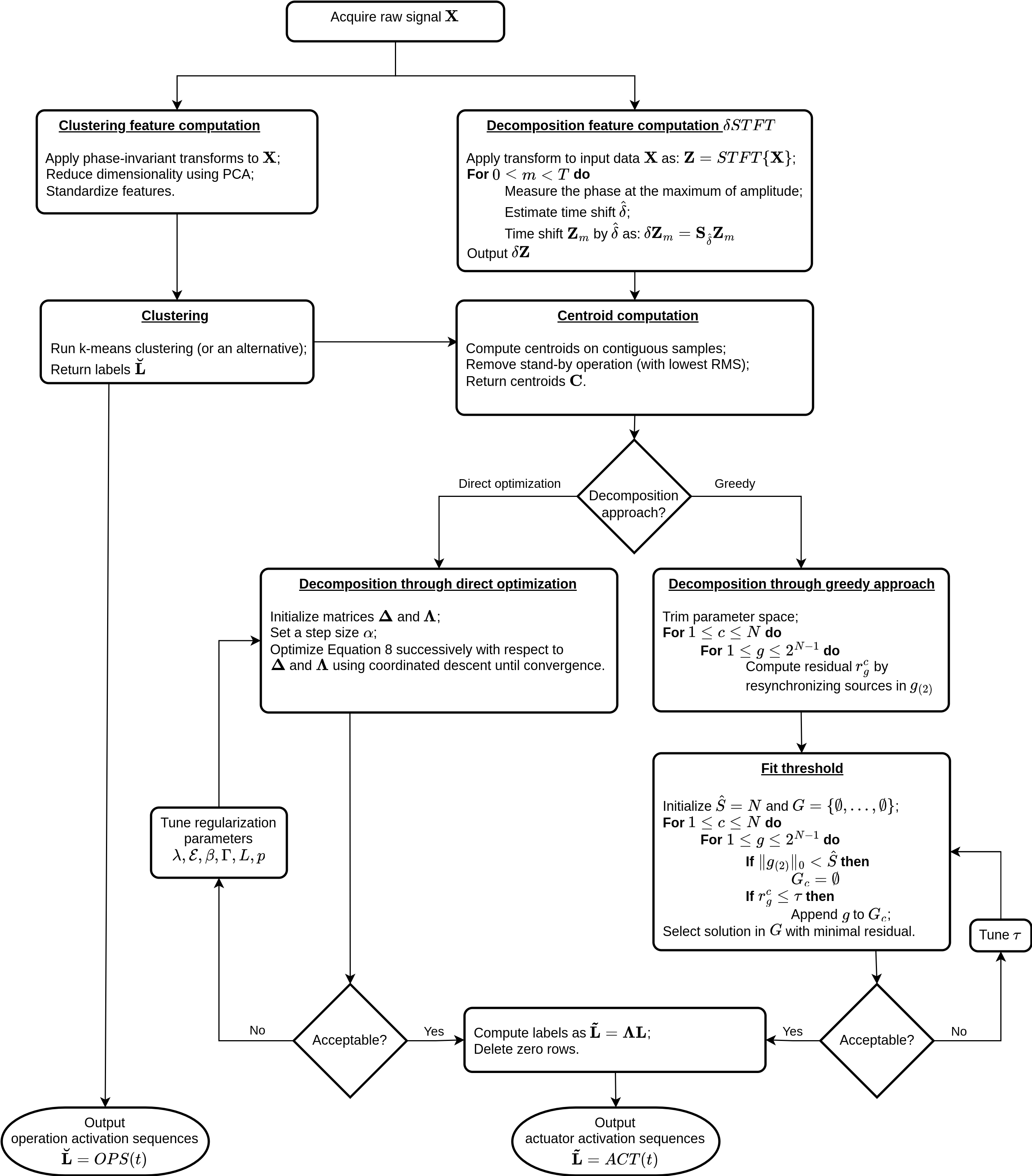}
    \caption{Flowchart of the $\mathbb{C}$SBMF for source activation sequence retrieval}
    \label{fig:flowchart}
\end{figure}

At last, the proposed $\mathbb{C}$SBMF algorithm to retrieve the activation sequences is presented in \prettyref{fig:flowchart}.

\subsection{Limitations}

There are cases in which machines operate at different regimes.
For instance, a motor operating at constant speed with different loads, monitored by an accelerometer, might produce the same signature (same power spectrum), varying only by a scaling factor.
Then recombining these sources during post-processing is straightforward.
Should the actuator produce different signatures under these regimes (different speeds in the previous example), recombination is not possible with the proposed technique.

Another practical issue is when distinct actuators produce the exact same signature.
If run concurrently, this case is no different from a single device operated at different regimes.
An indetermination hence remains between both cases.
In the greedy algorithm, multiplicity can be taken into account by expressing the parameter space in base $b$, with $b$ the maximum multiplicity, instead of base $2$.
The final complexity of this algorithm is $\mathcal{O}(N b^{N})$, times the optimizer's complexity as regards residual calculation.
This remains acceptable for monitoring small dedicated systems.

\section{Results}
\label{sec:results}

\subsection{Numerical experiments}
\label{sec:results_num}

We verify our method against synthetic signals to begin with. 
A representative scenario was selected here among our numerical experiments.
A piecewise stationary univariate signal $x$ is produced as the sequence of all possible sums of sources from an alphabet $\mathcal{C}$, containing a square wave $a$ (frequency $70Hz$, amplitude $1u$, zero-centered), a triangle wave $b$ (frequency $50Hz$, amplitude $1u$, zero-centered) and a sine wave $c$ (frequency $50Hz$, amplitude $2u$, zero-centered).
Signal is supplemented with a zero-mean Gaussian noise $w(t)$ with standard deviation $\sigma = 0.1 u$.

\begin{figure}[h!]
  \centering
  \includegraphics[width=1.\linewidth]{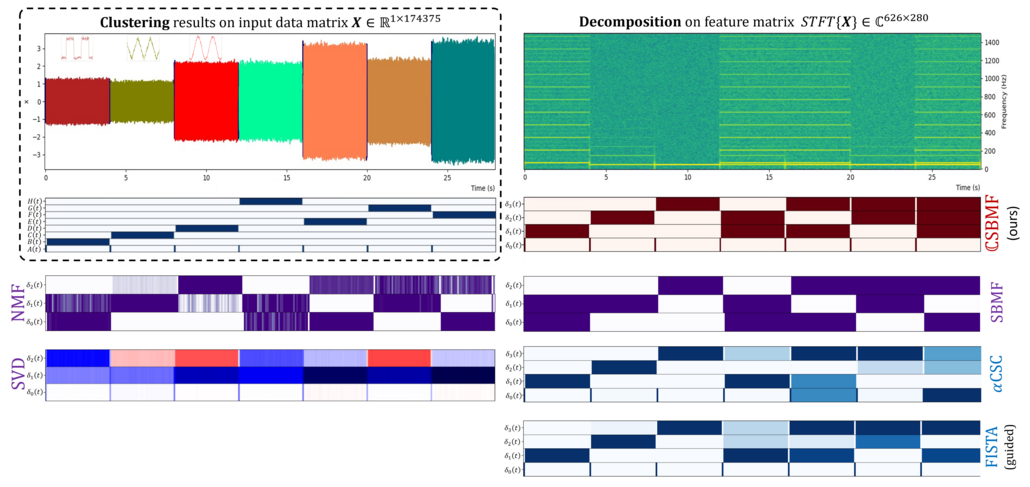}
  \caption{Decomposition of a synthetic signal ($70Hz$ square, $50Hz$ triangle and $50Hz$ sine waves, and their combinations). Six decomposition methods are presented: the $\mathbb{C}$SBMF (red), dictionary learning techniques (purple) and sparse coding (blue). Non-binary labels lie in [0, 1] (scale is dictionary-dependent otherwise).}
  \label{fig:result_decomposition_synthetic}
\end{figure}

In order to shed light on practical limitations of existing methods, we compare qualitatively the $\mathbb{C}$SMBF to traditional and state-of-the-art techniques tackling similar problems.
The results are presented in \prettyref{fig:result_decomposition_synthetic}.
The proposed benchmark comprehends semi-binary matrix factorization (SBMF) \cite{sorensen2022overlapping}, alpha-stable convolutional sparse coding ($\alpha$CSC) \cite{jas2017learning, dupre2018multivariate}, as well as NMF.
Sparse coding steps have been performed using the Fast Iterative Shrinkage-Thresholding Algorithm (FISTA) \cite{beck2009fast}.

The SBMF is limited to the study of real-valued signals.
Less ambiguous than NMF, it captures \say{the direct sum (as opposed to the average) of community activities} \cite{sorensen2022overlapping}.
Albeit similar to the proposed $\mathbb{C}$SBMF, it does not rely on assumption \prettyref{P:11} to build the dictionary and rather uses an SVD-based initialization.
It is hence better at capturing intrinsic characteristics, yet concerns remain as to the validity of the result and its interpretation, as will testify the decomposition in \prettyref{fig:result_decomposition_synthetic}.
Applied to spectrogram data, the SBMF accurately predicted the presence of the sine wave, but difficulties subsist in differentiating the triangle and square waves.
We suspect this behavior is caused by noise and spectrogram-induced nonlinearity.
To put the emphasis on the nonlinear aspect, the decomposition referred to as \textit{guided FISTA} in our experiment uses the optimal dictionary directly (power spectral density of each wave).
Projecting the spectrogram onto it, the triangle wave remains poorly identified.

In our investigations, the $\alpha$CSC, applied to the time series directly, was found to excel at retrieving temporal patterns.
By taking the DFT of these patterns to build the dictionary and projecting the STFT onto them, linearity is preserved.
While properly identifying the waves alone, resynchronization is absent from this process, and indeed this method fails to retrieve the combinations.

In comparison to these techniques, the $\mathbb{C}$SBMF reliably finds meaningful centroids using clustering, and effectively recovers the activation sequences.
In our experiments, source resynchronization allowed to lower the residuals of the desired decompositions by at least two orders of magnitude with respect to their countepart computed using the modulus of the atoms.
The performance of the proposed method is tied to the effectiveness of the clustering as well as the averaging process, which is affected by transients, outliers and noise distributions.
We therefore validate the $\mathbb{C}$SBMF on real-world signals.

\subsection{Experimental results}
\label{sec:results_exp}

\begin{figure}[h!]
  \centering
  \includegraphics[width=1.\linewidth]{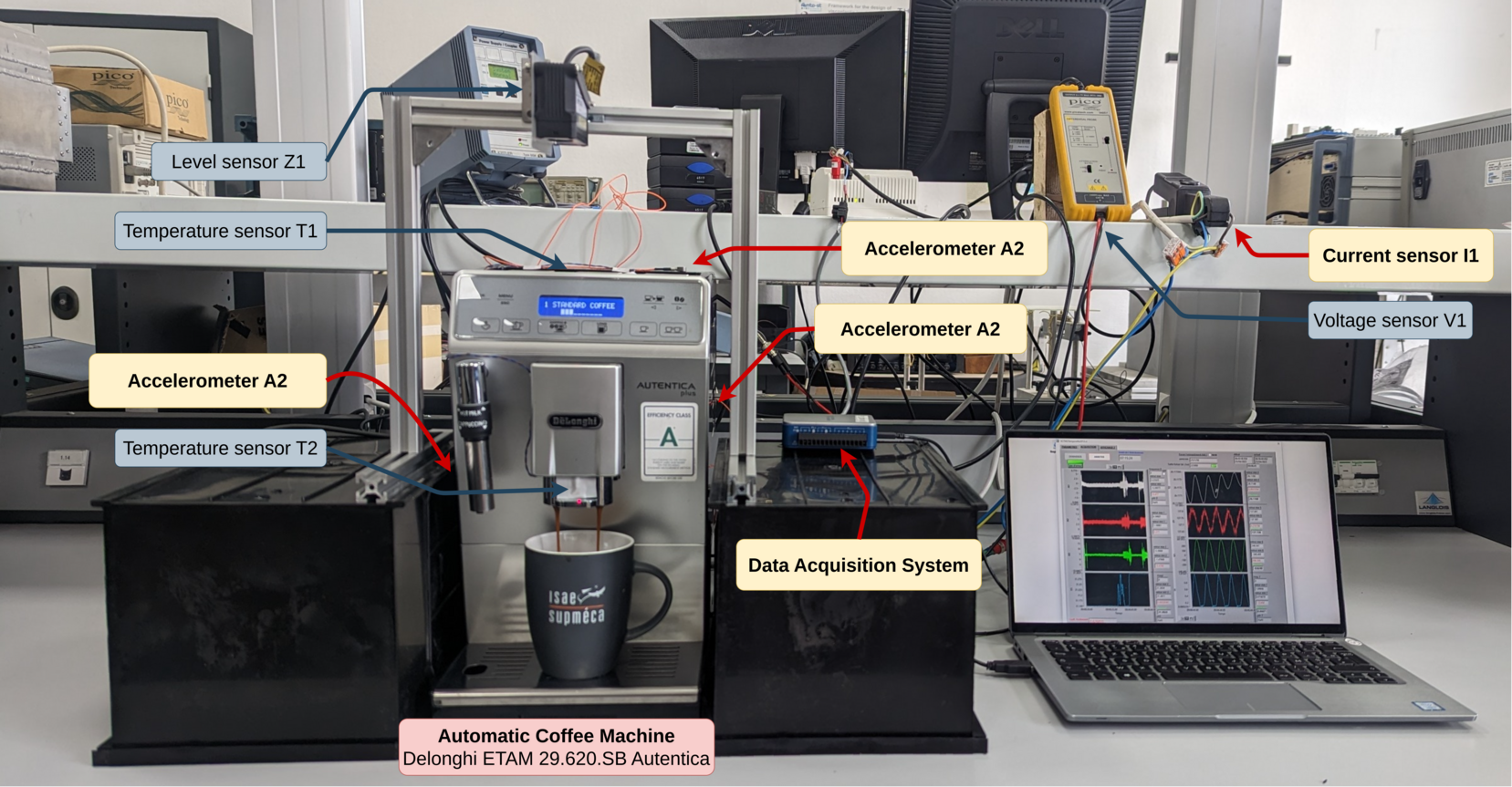}
  \caption{Experimental setup of the CAFFEINE dataset \cite{delabeye_romain_2023_8351431} using the NI USB-6003 data acquisition system (16bit, 8channels, 6.25kHz)}
  \label{fig:CAFFEINEsetup}
\end{figure}

As a condensed version of an industrial system, we validate the $\mathbb{C}$SBMF on the CAFFEINE dataset \cite{delabeye_romain_2023_8351431} presented in \prettyref{fig:CAFFEINEsetup}, and more specifically on the current and vibration signals.
This use case consists in an automated coffee machine, made of four multiphysical actuators: one \textit{heating coil}, one vibration \textit{pump}, one \textit{infuser} (motor with a worm gear to displace the infuser) and one \textit{grinder} (motor with an epicyclic gearing to grind coffee beans).
The relevance of such a system for industrial applications was shown in \cite{delabeye2022scalable}.

\begin{figure}[H]
  \centering
  \includegraphics[width=1.\linewidth]{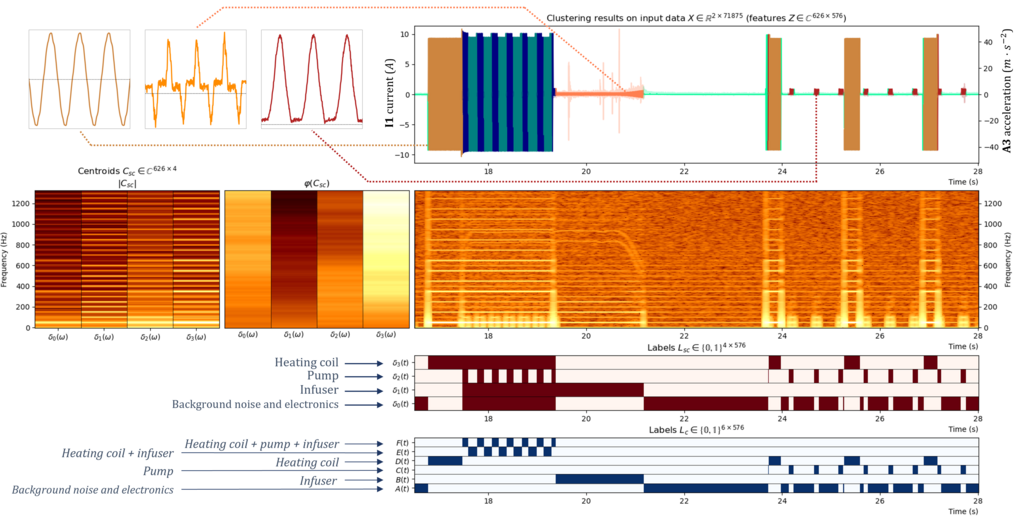}
  \caption{Decomposition of the current signal, trial 42 of the CAFFEINE dataset}
  \label{fig:result_decomposition_CAFFEINEcurrent}
\end{figure}

The results of the $\mathbb{C}$SBMF applied to current sensor data are presented in \prettyref{fig:result_decomposition_CAFFEINEcurrent}.
The residual threshold was set to $\tau = 0.0315$.
We observed that this hyperparameter was more sensitive as the sources' scales were unbalanced.
Here, the heating coil consumes 65 times more than the infuser, and the infuser's RMS is only twice that of the background noise.
On this dataset, the activation sequences of the three actuators are well recovered despite slight nonstationarities and noise.
In order to push the boundary of our method, centroids of background noise and low consumption electronics were kept.

\begin{figure}[H]
  \centering
  \includegraphics[width=1.\linewidth]{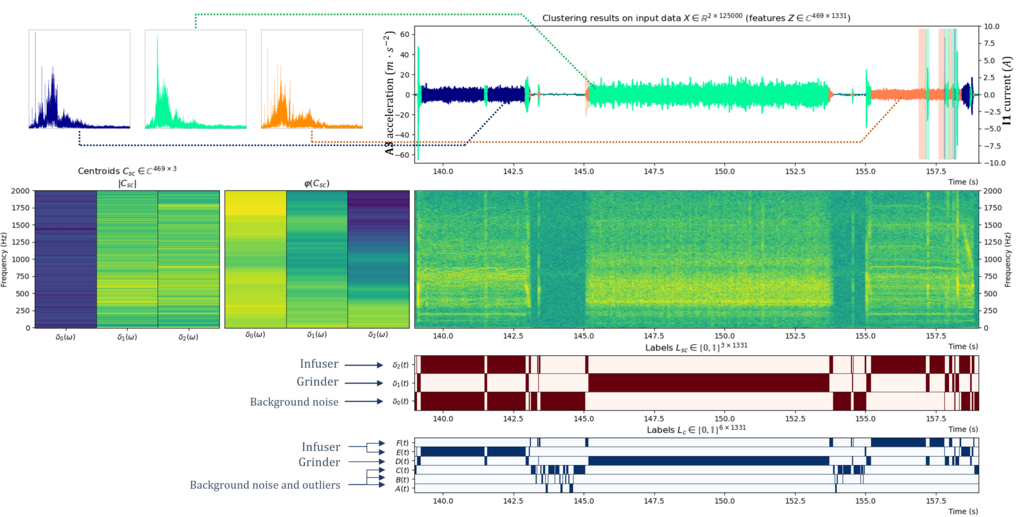}
  \caption{Decomposition of the accelerometer signal, trial 42 of the CAFFEINE dataset}
  \label{fig:result_decomposition_CAFFEINEaccelerometer}
\end{figure}

In \prettyref{fig:result_decomposition_CAFFEINEaccelerometer}, we apply the $\mathbb{C}$SBMF on accelerometer data.
These components produce extremely noisy (grinder) and partially nonstationary (infuser) signals.
This behavior breaks the quasi-stationarity assumption, which translates into inaccurate centroid estimation and leads to incorrect classification.
In spite of these challenging conditions, an alternate use for the $\mathbb{C}$SBMF is to recombine wrongfully clustered centroids.
Indeed, as an intra-cluster variance minimization algorithm, $k-means$ often performs better when overestimating the number of clusters.
This isolates outliers.
In this case, the outliers and the actuators activated in different operations share similarities, making it possible to regroup them in a meaningful fashion.

\section{Discussion}
\label{sec:discussion}

In this work, we highlight the importance of keeping data's extensive property to recover the activation sequences.
Be it through convolutions in the time domain \cite{zhou2016detection, dupre2018multivariate}, or phase resynchronization in the TF domain as proposed here, we qualify the matrix factorization as \textit{exact} insofar as most of the signal information has been preserved.
Although a phase-preserving decomposition is interesting, (i) other sources of uncertainty remain, noise and transients in particular, and (ii) the approach is inherently subject to the curse of dimensionality.
Indeed, retaining the properties of the DFT forces to represent data in a high dimensional space, thus limiting the discrimination between centroids or samples.
For this reason, using nonlinear transforms, or even simple standardization, to learn a more discriminating manifold would be beneficial \cite{balshaw2022importance}.
This could also help lifting assumption \prettyref{P:10} to tackle cases where the subsystems are coupled, e.g., in series association.

The $\mathbb{C}$SBMF heavily relies on clustering to build a dictionary.
Whilst this is a debatable choice for compressed sensing, its relevance to source identification is made clear in this paper.
This approach is particularly interesting as regards (i) computational complexity (decomposing centroids instead of samples, $N \ll T$), (ii) resilience to noise thanks to the averaging process, and (iii) estimation of the number of sources.
Clustering also constitutes the method's Achilles' heel, as the decomposition is as accurate as the clustering technique is.
Centroids are also subject to transient-originated outliers, and their misestimation is detrimental to the factorization process.
Robust kernel smoothing can be used to compute outlier-free centroids though \cite{humbert2022robust}.

At last, the $\mathbb{C}$SBMF overlooked the case where a system is operated at different regimes.
Periodic regularization functions could be considered to extend the representation's domain to $\mathbb{Z}$ ($\mathbb{N}$, ideally) \cite{naumov2018periodic}, a common practice in quantization neural networks.
Should the signature shift however smoothly from a regime to another, matrix decomposition is no longer appropriate.
Graph neural networks (GNN) excel at this type of task, therefore making them good candidate architectures to learn an embedding in which sources can be easily separated \cite{shchur2019overlapping}.
Tracking could also be used to that effect, learning trajectories and their principal characteristics instead of centroids \cite{delabeye2023sequential}.
Overall, the rationale is that a single vector may prove insufficient to represent a source in multiple configurations.

\section{Conclusions and perspective}
\label{sec:conclusions}

In this paper, the $\mathbb{C}$SBMF method was proposed to recover source activation sequences in mixed stationary periodic signals.
This study highlighted limitations in traditional methods in challenging conditions, where the sources may be correlated and their number difficult to estimate.
A formulation to this semi-binary decomposition was proposed as a phase-preserving bi-variate optimization problem.
Although direct solving proved tedious and convergence could not be guaranteed, the proposed greedy algorithm stems from a meticulous study of this formulation.
A novel operator, coined $\delta STFT$, was introduced in an effort to extract meaningful centroids in the complex plane, thus keeping the Fourier transform's linearity.
Additionally, a phase resynchronization mechanism allowed to express centroids with respect to others, and thus find a minimal basis in which data can be reconstructed.
Finally, due to spurious minimizers --- both in dictionary and representation learning --- jeopardizing the optimization process, and building up on the fact that only a tiny proportion of the representation's parameter space is actually relevant, a greedy algorithm was designed.

This work paves the way for interesting prospects.
As a trade-off between effective variants of the NMF and the proposed $\mathbb{C}$SBMF, an efficient parameter space reduction and search could be proposed by using a phase-invariant space as proxy before resynchronization, without loss of generalization.
Future work will also aim for direct solving of \prettyref{eq:main} using scalable optimization techniques and building up a lower dimensional dictionary.
That is, in light of this work, we suspect a more discriminating dictionary could be learnt appropriately.
Although the absence of training is interesting, this limits the potential for performance gains.
We believe deep-learning-based underdetermined blind source separation techniques will benefit from the findings presented in this paper, by avoiding phase-shift-induced pitfalls in particular.

Number of industrial use cases contain transient, non-stationary and disturbed signals.
At present, our method is very sensitive to these non-stationarities, although experimental validation has shown it performed well under mild quasi-stationary conditions in a representative use case.
Means to tackle greater levels of non-stationarity and outliers in source signals will hence be investigated.

\section*{Acknowledgments}

This work was funded by the EnerMan project from the European Union’s Horizon 2020 research and innovation programme under grant agreement No 958478.
The authors would also like to thank Christophe Ben Brahim for their support in designing the experimental setup.



\appendix

\section{Detailed study of the $\mathbb{C}$SBMF formulation}
\label{apx:opt}

This section details the mechanisms at play in the $\mathbb{C}$SBMF formulation (\prettyref{eq:main}).
In particular, we highlight the existence of a global minimum, or at least we show through a generic example that the minimizers are not equivalent to one another under suitable regularization conditions.
Inequalities for some of the regularization parameters are provided.
This constrains the hyperparameters to tune and guarantees the validity of the expected properties of each penalty.

A simple sequence \say{$a-b-ab$} is considered as a synthetic use case, where $a$ and $b$ are normalized $T$-periodic triangle and square waves respectively.
Operation $ab$ is the sum of $a$ and $b$, shifted by $T/4$ and $T/3$ respectively.
$F$ denotes the least squares functional of \prettyref{eq:main} and $J$ is the complete cost function including all regularizers.

\subsection{Tikhonov regularization on the time shifts}

\begin{figure}
\centering
\begin{minipage}{.5\textwidth}
  \centering
  \includegraphics[width=1.\linewidth]{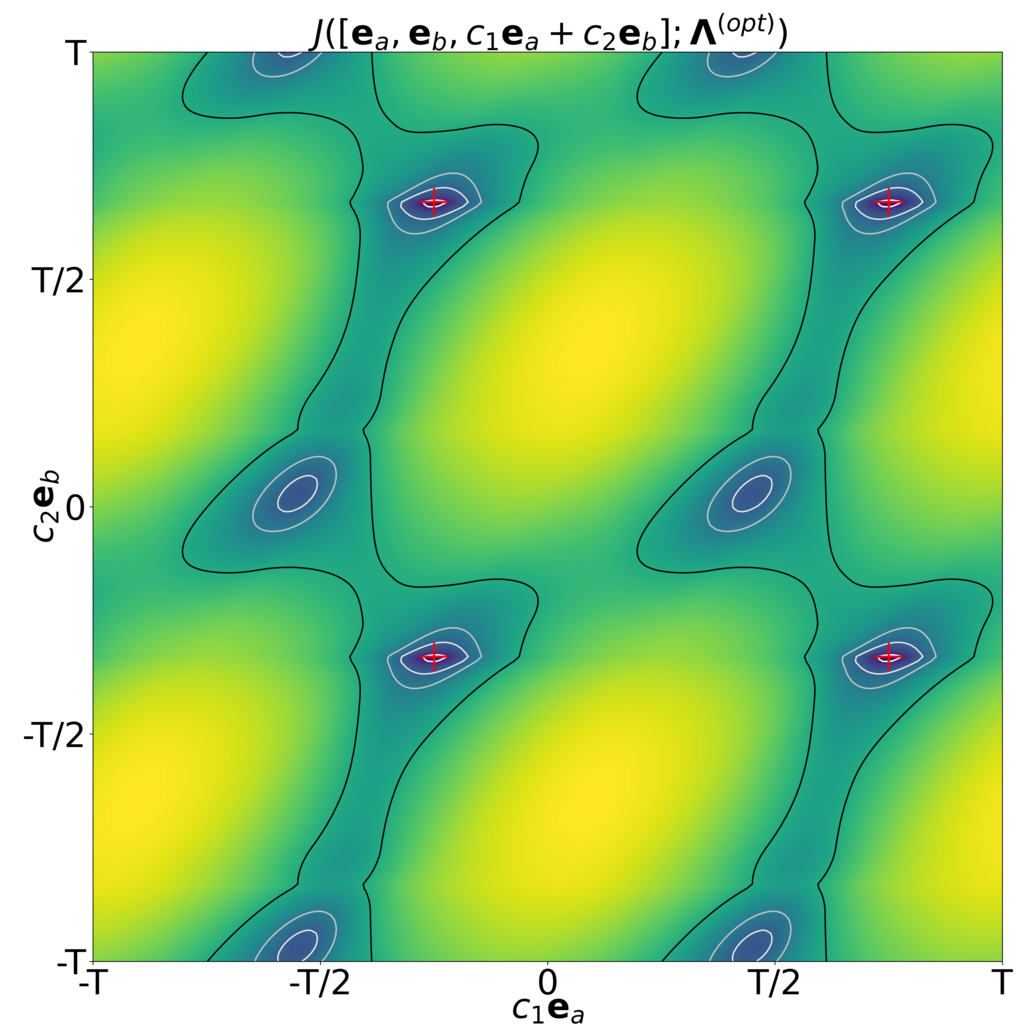}
  \caption{No regularization}
  \label{fig:opt_resync_noregul}
\end{minipage}%
\begin{minipage}{.5\textwidth}
  \centering
  \includegraphics[width=1.\linewidth]{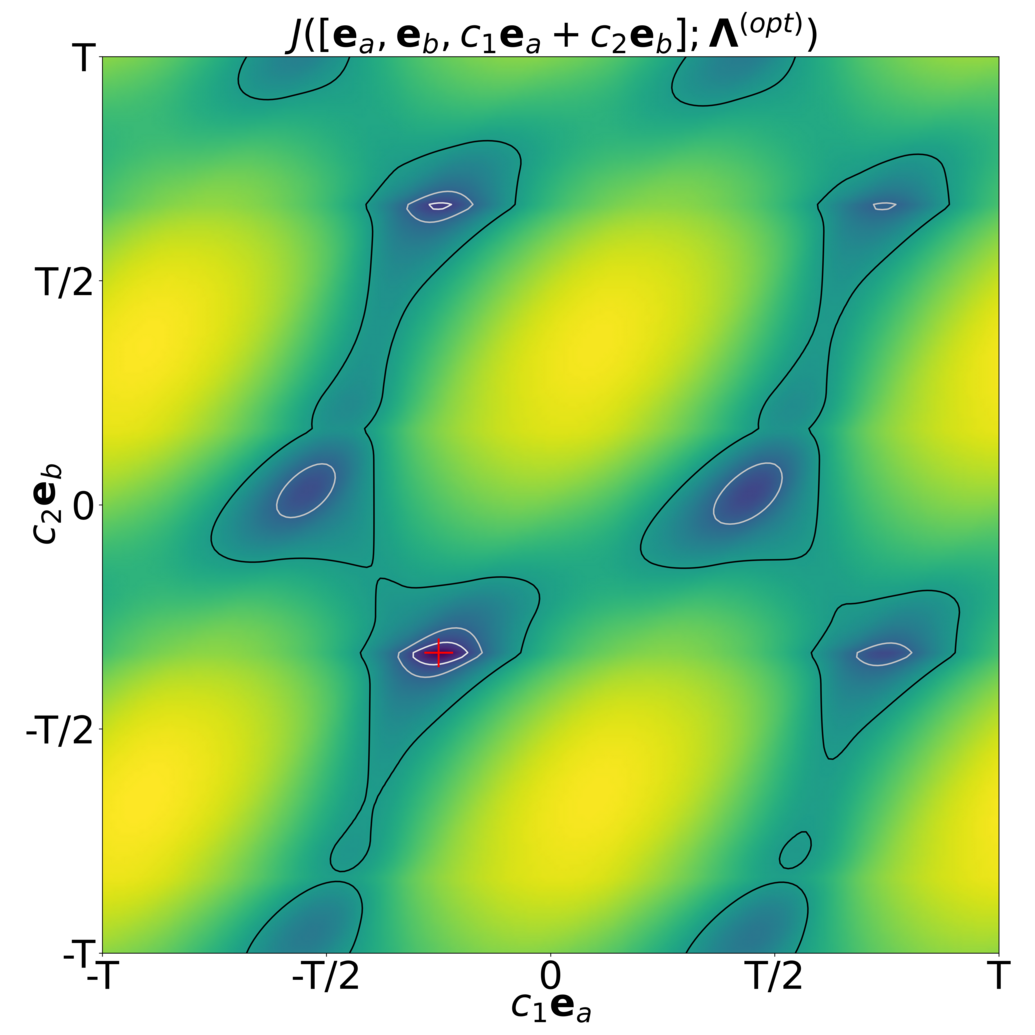}
  \caption{Tikhonov regularization}
  \label{fig:opt_resync_tikhonov}
\end{minipage}
\caption{Minimizers of $F(\bm{\Delta}_{ab}, \bm{\Lambda}_{ab}^{opt})$ (marked in red, with level curves)}
\label{fig:opt_resync}
\end{figure}

Due to periodicity, there is an infinite number of minimizers enabling the resynchronization of each atom $i$ to reconstruct another atom $c$ using time shifts $\Delta_c^i \in \mathbb{R}$.
Indeed, the time shift operator is periodic, hence so is the functional $F(\bm{\Delta}, \bm{\Lambda})$.
Each time shift is applied to a single atom though, $F(\bm{\Delta}, \bm{\Lambda})$ is hence periodic of period $\hat{T}^{(i)}$ along each dimension $\Delta_c^i$.
Given the desired representation $\bm{\Lambda}_{ab}^{(opt)} = [\bm{e}_a, \bm{e}_b, \bm{e}_{a} + \bm{e}_b]$, \prettyref{fig:opt_resync} shows the presence of multiple minimizers ($T$-periodic), with $\bm{e}_a$, $\bm{e}_b$, $\bm{e}_{ab}$ the canonical vectors.
Uniqueness of the solution in $\bm{\Delta}$ is obtained through Tikhonov regularization ($\Gamma = 10^{-5}$).

\subsubsection{Regularization of the representation}

Three types of regularizations are implemented for (i) sparsity, (ii) binarity, and (iii) consistency as regards the energy profiles of the decompositions.
Diverse experiments are conducted to study the effect of these regularizers.
The following parameterization is used to represent the cost function in two dimensions: 
$\bm{\Lambda}_{ab} = c_1 \bm{e}_a + c_2 \bm{e}_b$ (\prettyref{eq:opt_exp_lambda_ab_a_b}), $\bm{\Lambda}_{ab} = (\frac{c_1 + c_2}{2}) \bm{e}_{ab}$ (\prettyref{eq:opt_exp_lambda_ab_ab}), $\bm{\Lambda}_b = c_1 \bm{e}_a + c_2 \bm{e}_{ab}$ (\prettyref{eq:opt_exp_lambda_b_ab_a}) and $\bm{\Lambda}_a = c_1 \bm{e}_{ab} + c_2 \bm{e}_b$ (\prettyref{eq:opt_exp_lambda_a_ab_b}).

\noindent\begin{minipage}[c]{.25\linewidth}
\begin{equation}\label{eq:opt_exp_lambda_ab_a_b}
    \bm{\Lambda} =
    \begin{bmatrix}
        1 & 0 & c_1\\
        0 & 1 & c_1\\
        0 & 0 & c_2
    \end{bmatrix}
\end{equation}
\end{minipage}%
\noindent\begin{minipage}[c]{.25\linewidth}
\begin{equation}\label{eq:opt_exp_lambda_ab_ab}
    \bm{\Lambda} =
    \begin{bmatrix}
        1 & 0 & 0\\
        0 & 1 & 0\\
        0 & 0 & \frac{c_1 + c_2}{2}
    \end{bmatrix}
\end{equation}
\end{minipage}%
\noindent\begin{minipage}[c]{.25\linewidth}
\begin{equation}\label{eq:opt_exp_lambda_b_ab_a}
    \bm{\Lambda} =
    \begin{bmatrix}
        1 & c_1 & 0\\
        0 & 0 & 0\\
        0 & c_2 & 1
    \end{bmatrix}
\end{equation}
\end{minipage}%
\begin{minipage}[c]{.25\linewidth}
\begin{equation}\label{eq:opt_exp_lambda_a_ab_b}
    \bm{\Lambda} =
    \begin{bmatrix}
        0 & 0 & 0\\
        c_2 & 1 & 0\\
        c_1 & 0 & 1
    \end{bmatrix}
\end{equation}
\end{minipage}
\vspace{\dp0}

From this point on, it is assumed that the optimal time shifts $\bm{\Delta}^{opt}$ have been reached.

\subsubsection{Sparsity regularization}

\begin{figure}
\centering
\begin{minipage}{.33\textwidth}
  \centering
  \includegraphics[width=1.\linewidth]{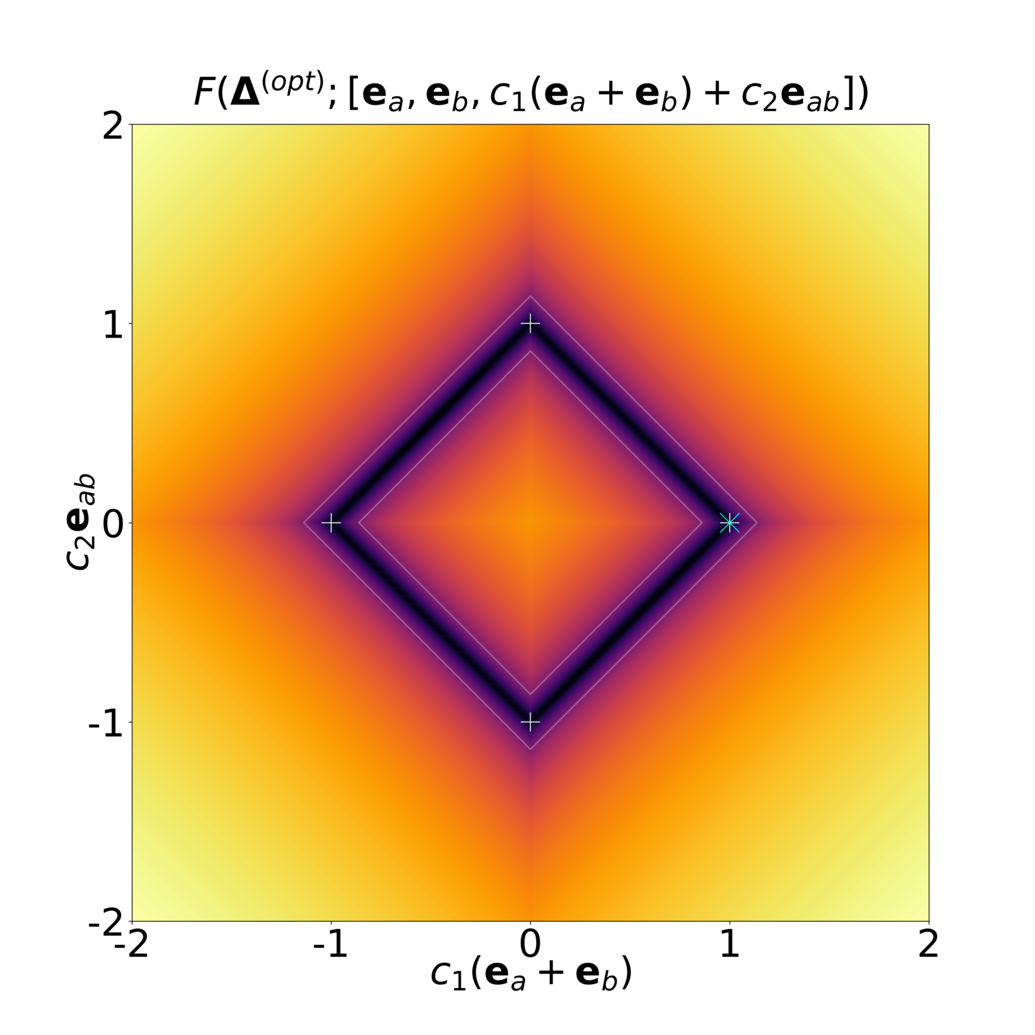}
  \caption{None}
  \label{fig:opt_sparse_noregul}
\end{minipage}%
\begin{minipage}{.33\textwidth}
  \centering
  \includegraphics[width=1.\linewidth]{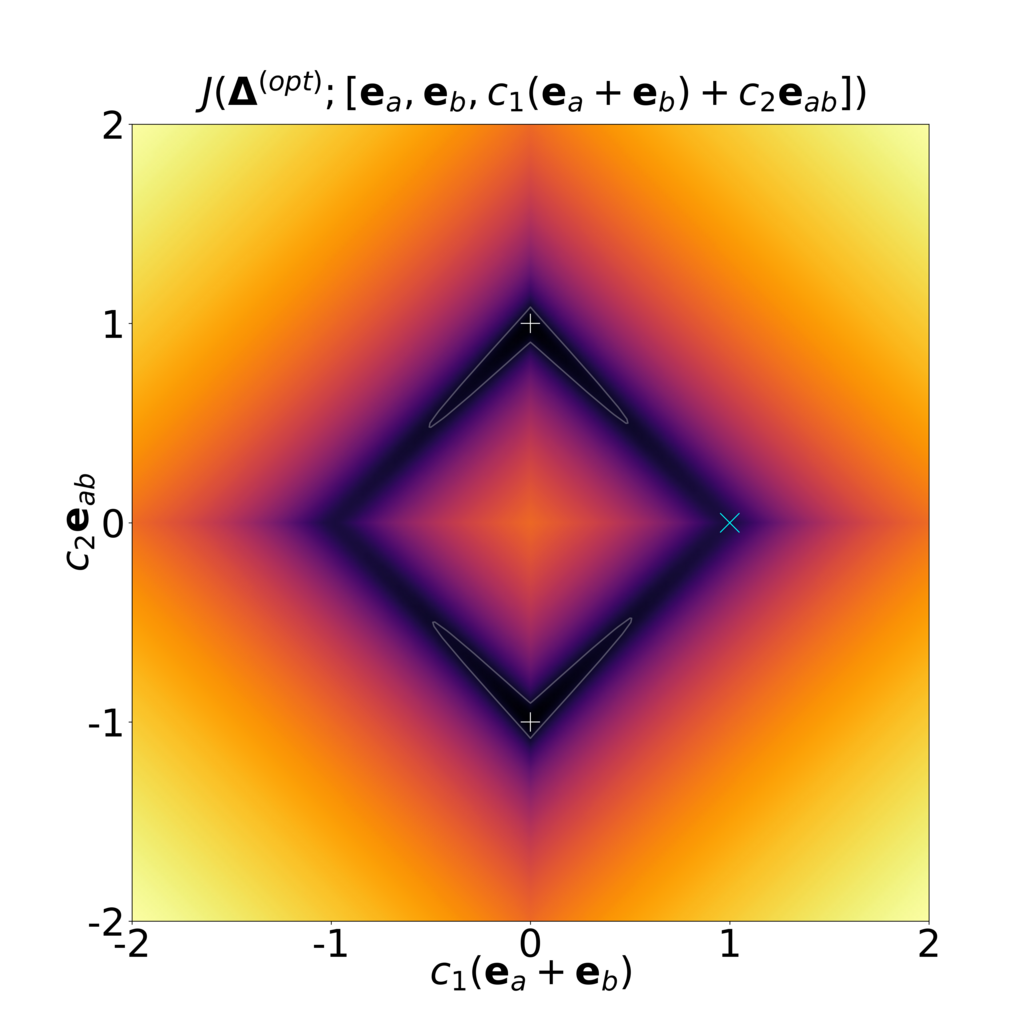}
  \caption{$\ell_p$}
  \label{fig:opt_sparse_col}
\end{minipage}%
\begin{minipage}{.33\textwidth}
  \centering
  \includegraphics[width=1.\linewidth]{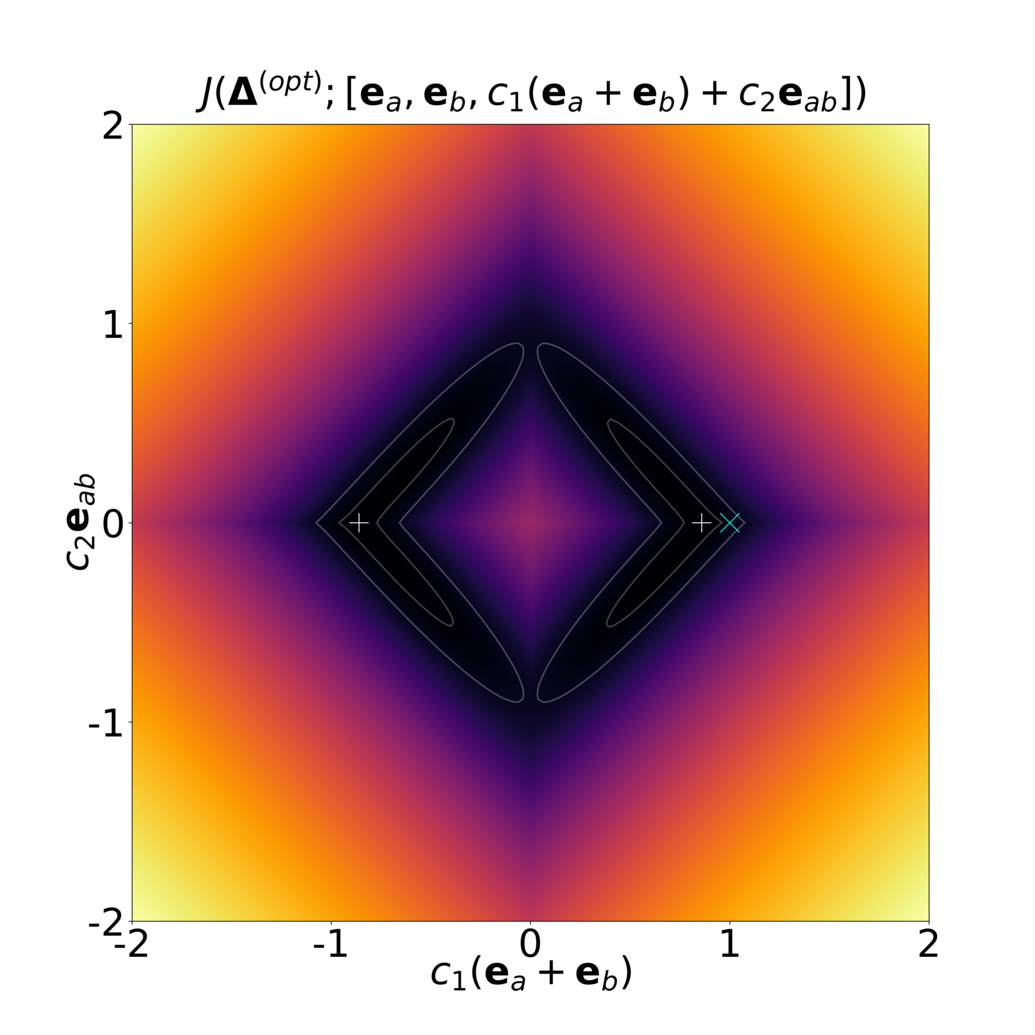}
  \caption{$\ell_p$ and $\ell_{2,p}$}
  \label{fig:opt_sparse_colrow}
\end{minipage}%
\caption{Minimizers of $J_{sparse}(\bm{\Delta}_{ab}^{(opt)}, \bm{\Lambda}_{ab})$ (marked in white, with level curves)}
\label{fig:opt_sparse}
\end{figure}

The functional $F(\bm{\Delta}^{(opt)}, \bm{\Lambda})$ includes a number of minima as regards $\bm{\Lambda}$.
Sparsity regularization usually consists in minimizing the $\ell_0$ norm of a vector or an estimator thereof.
This produces two noticeable effects: (i) it allows to reconstruct a sample with a minimal number of relevant atoms, (ii) negligible coefficients tend to zero.
A difficulty in using a dictionary derived from clustering lies in the fact that the desired decomposition maximizes the number of relevant components instead of minimizing it.
For this reason, the trivial solution $\bm{\Lambda} = \bm{I}$ always exists and it is the easiest minimum to find.
To remedy this limitation, classic $\ell_p$ regularization on the columns of $\bm{\Lambda}$ is coupled with $\ell_{2,p}$ penalization on $\bm{\Lambda}^T$, with $p \leq 1$.
The latter estimates the number of nonzero rows in $\bm{\Lambda}$ (number of sources).
This mechanism is applied to the synthetic use case in \prettyref{fig:opt_sparse}, where $J_{sparse}$ denotes the cost function including the functional as well as the column-wise $\ell_p$ and row-wise $\ell_{2, p}$ sparsity promoting penalties.

As illustrated in \prettyref{fig:opt_sparse_noregul}, an infinite number of minimizers are connected through valleys as regards the representation in the absence of regularization.
That is, the reconstruction is a weighted sum of several admissible combinations.
When an element of $\bm{\Lambda}$ is negative, the corresponding atom may be flipped as a result of resynchronization (worst case).
\prettyref{fig:opt_sparse_col} shows the effect of column-wise sparsity regularization --- pushing $\bm{\Lambda}$ towards the identity ---, whereas \prettyref{fig:opt_sparse_colrow} is the result of both column- and row-wise sparsity penalization, --- minimizing the number of nonzero rows while penalizing negligible coefficients ---.
The scale is lost in the process though.
The discrepancy may be observed on \prettyref{fig:opt_sparse_colrow} between the minimum marked in white and the expected solution in cyan.

The tradeoff between sparse penalties is found by prioritizing the estimation of the number of sources:

\begin{equation}\label{eq:Gamma_lowerbound_init}
    \sum_{c=1}^N \Bigg ( \lambda + \frac{\mathcal{E}}{\lVert \bm{C}_c \rVert_2^2} \Bigg ) \lVert \bm{\Lambda}_c \rVert_p < L \lVert \bm{\Lambda}^T \rVert_{2, p}
\end{equation}

which, by maximizing the left hand side (maximum decomposition $\mathbbm{1}_{N-1}$, given the least energetic centroid $\mu_{min} = \displaystyle\min_{c} \lVert \bm{C}_c \rVert_2^2$) and minimizing the right hand side (one source), yields:

\begin{equation}\label{eq:Gamma_lowerbound}
    L > (\lambda + \frac{\mathcal{E}}{\mu_{min}}) N \lVert \mathbbm{1}_{N-1} \rVert_p
\end{equation}

\subsubsection{Binary regularization}

\begin{figure}[H]
\centering
\begin{minipage}{.5\textwidth}
  \centering
  \includegraphics[width=1.\linewidth]{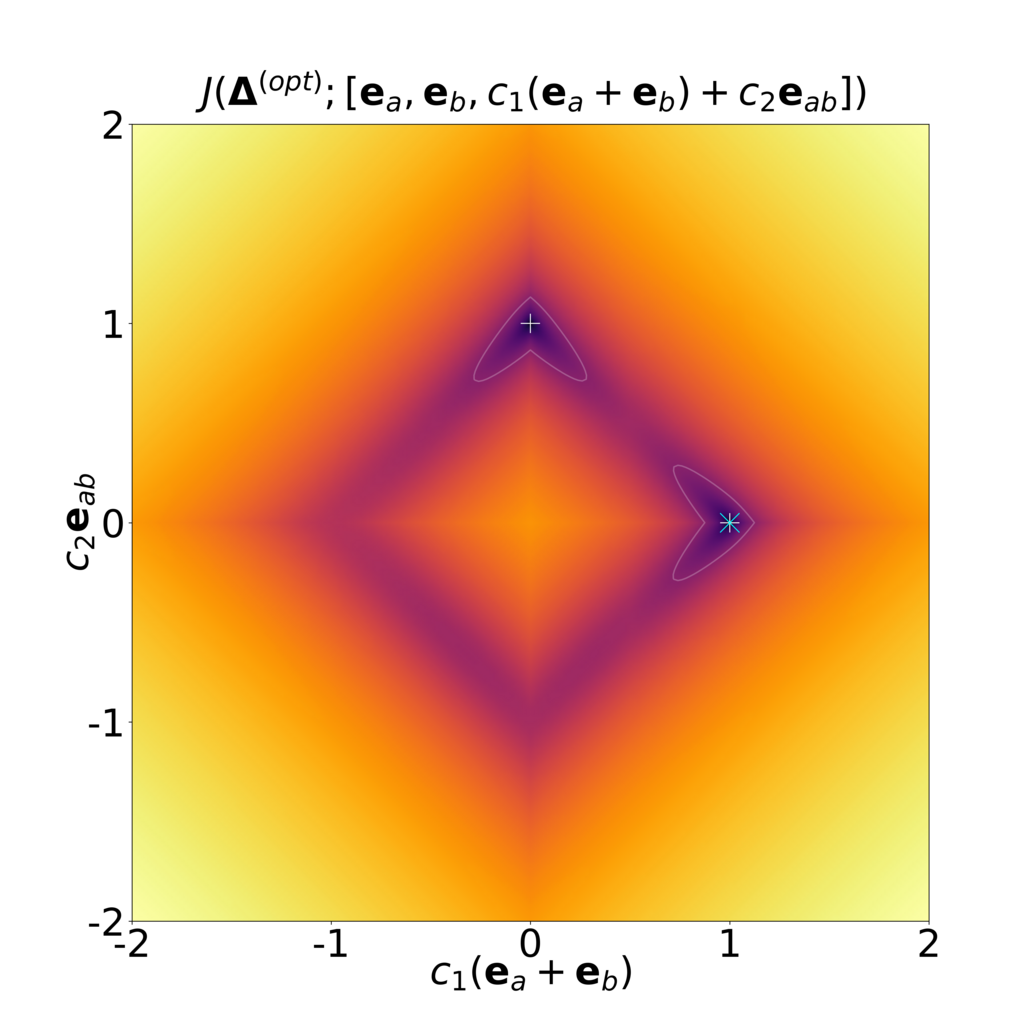}
  \caption{Binary regularization only}
  \label{fig:opt_binary_noregul}
\end{minipage}%
\begin{minipage}{.5\textwidth}
  \centering
  \includegraphics[width=1.\linewidth]{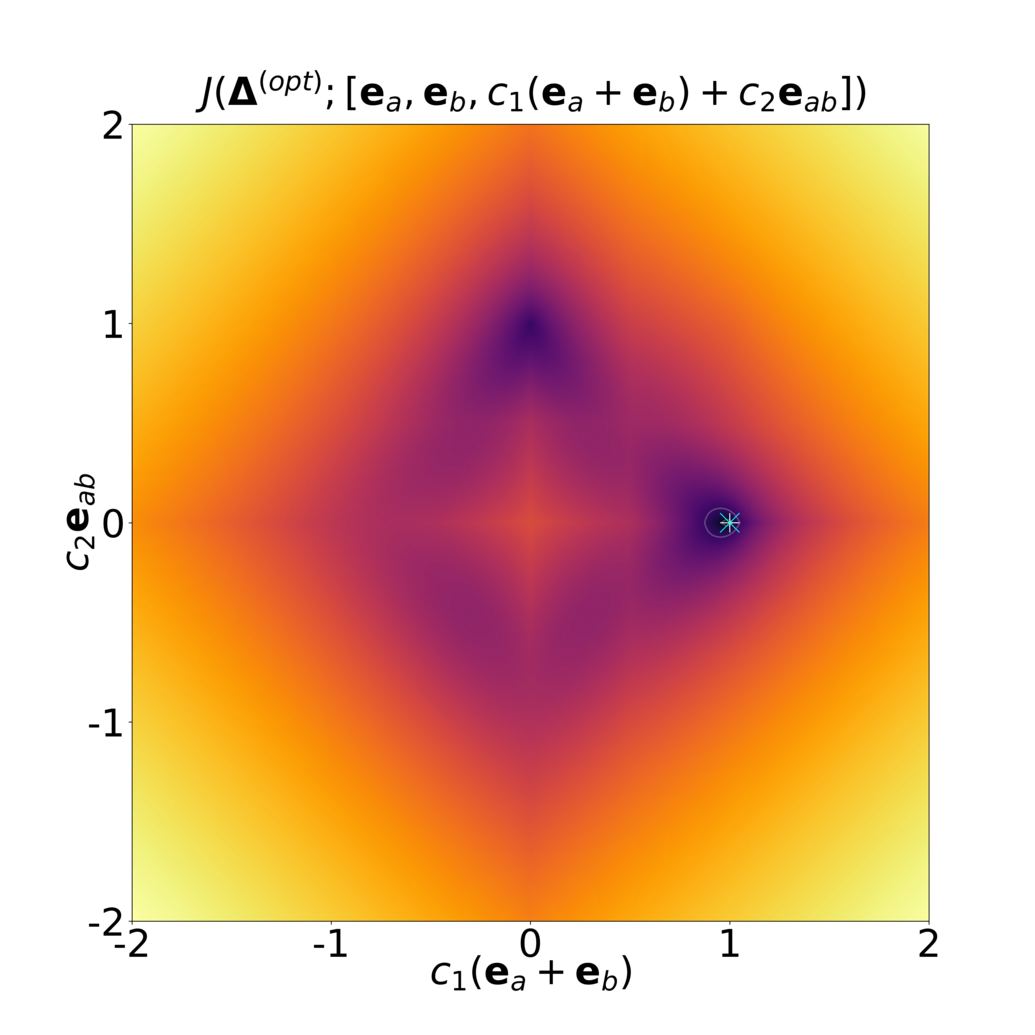}
  \caption{Binary and sparse penalties}
  \label{fig:opt_binary_sparse}
\end{minipage}
\caption{Minimizers of $J(\bm{\Delta}_{ab}^{(opt)}, \bm{\Lambda}_{ab})$ (marked in red, with level curves)}
\label{fig:opt_binary}
\end{figure}

Binary regularization as it is referred to in this paper is a special case of quantization.
The penalty is zero when $\bm{\Lambda}$ is binary.
The expected behavior of this regularizer is to displace the desired minimum towards binary locations.
This effect appears in \prettyref{fig:opt_binary}.

Worth noting, the least penalized location is $\bm{\Lambda}=0$.
This location corresponds to the sum of the atoms' squared norms (all penalties are null).
Hence care must be taken to have a higher cost at the center than at every minimizer's location $\bm{\Lambda}^{(opt)}$.
This condition binds $\lambda$, $L$, $\mathcal{E}$ and $\Gamma$ together as $J(\bm{\Delta}^{(opt)}, 0) < J(\bm{\Delta}^{(opt)}, \bm{\Lambda}^{(opt)})$, where $J$ is the complete cost function, in which $\mathcal{B}_2$ can be considered null.

\subsubsection{Regularization for iso-cardinality combinations}

\begin{figure}[H]
\centering
\begin{minipage}{.5\textwidth}
  \centering
  \includegraphics[width=1.\linewidth]{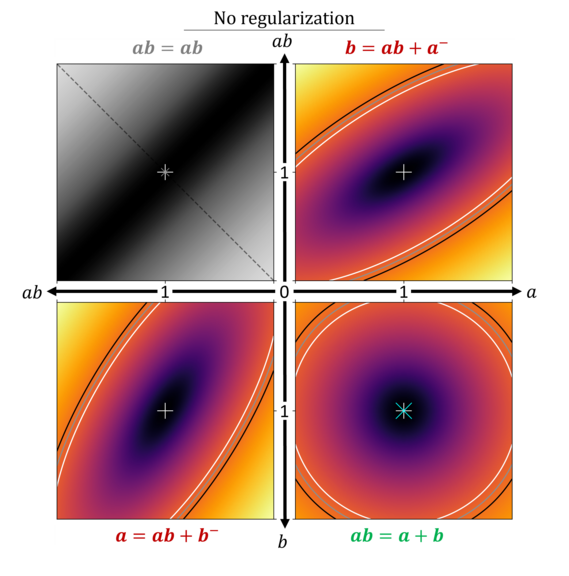}
  \caption{No regularization}
  \label{fig:opt_full_noregul}
\end{minipage}%
\begin{minipage}{.5\textwidth}
  \centering
  \includegraphics[width=1.\linewidth]{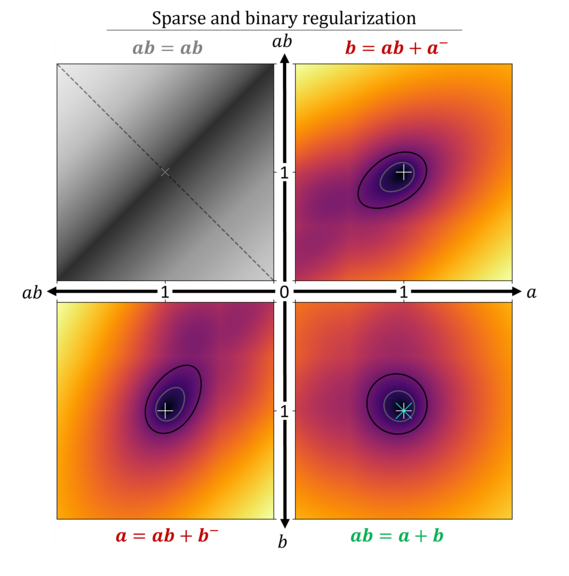}
  \caption{Binary and sparse only}
  \label{fig:opt_full_notriangle_sparse_binary}
\end{minipage}
\caption{Minimizers of $F(\bm{\Delta}^{(opt)}, \bm{\Lambda})$ (marked in red, with level curves)}
\label{fig:opt_full}
\end{figure}

At last, differentiating between combinations with the same cardinality involving the same atoms is problem-specific.
Geometrically, these solutions are equivalent, albeit the penalty on their respective time shifts may differ.
The regularization term $\mathcal{T}$ defined in \prettyref{eq:def_regularizer_triangle} is proposed to tell these solutions apart, by prioritizing solutions in which the atom to reconstruct has the highest energy level.
That is, for two minimizers $c$ and $j$, if \mbox{$\lVert \bm{\Lambda}^{(c)} \rVert_p = \lVert \bm{\Lambda}^{(j)} \rVert_p$} and \mbox{$\lVert \bm{C}_c \rVert_2^2 > \lVert \bm{C}_j \rVert_2^2$}, then coefficients $\mathcal{E}$ and $\Gamma$ must be such that $J(\bm{\Delta}^{(c)}, \bm{\Lambda}^{(c)}) < J(\bm{\Delta}^{(j)}, \bm{\Lambda}^{(j)})$, where $\mathcal{L}_{col}, \mathcal{L}_{row}$ cancel out and $\mathcal{B}_2, F$ are zero.
As shown on \ref{fig:opt_full_notriangle_sparse_binary}, where all regularizers are present but $\mathcal{T}$, without further consideration iso-cardinality combinations correspond to equivalent minima.

\begin{figure}[H]
  \centering
  \includegraphics[width=.8\linewidth]{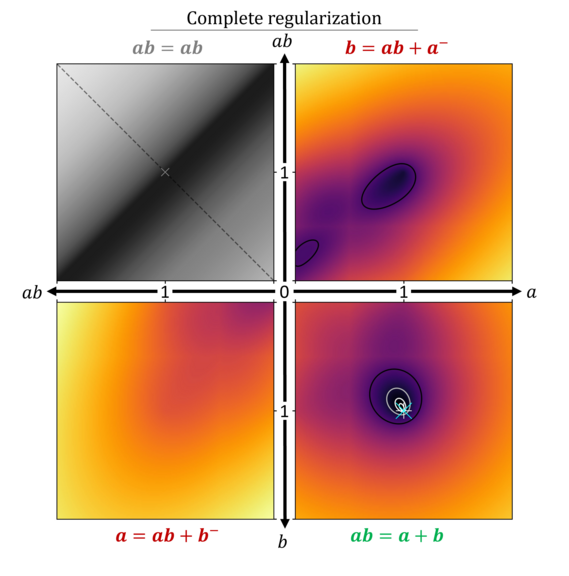}
  \caption{Complete regularization (\prettyref{eq:main})}
  \label{fig:opt_full_allregul}
\end{figure}

Finally, applying all penalties, a single minimum remains, as illustrated in \prettyref{fig:opt_full_allregul}.
The values used to regularize the problem in this example are as follows: $\Gamma = 10^{-5}$, $\lambda = 0.0225$, $L = 3.43$, $\beta = 0.7$, $\mathcal{E}=0.1$ and $p=0.9$.

 \bibliographystyle{Template/elsarticle-num} 
 \bibliography{refs.bib}

\end{document}